%% file: acl_long_en.tex
\documentclass[11pt]{article}

\usepackage{acl}
\usepackage{times}
\usepackage{latexsym}
\usepackage[T1]{fontenc}
\usepackage{microtype}
\usepackage{inconsolata}
\usepackage{graphicx}
\usepackage{subcaption}
\usepackage{booktabs}
\usepackage{multirow}
\usepackage{amsmath}
\usepackage{xcolor}
\usepackage{iftex}
\usepackage{tikz}

\ifPDFTeX
  \errmessage{This draft requires XeLaTeX. Please compile with xelatex.}
\fi

\usepackage{fontspec}
\usepackage{xeCJK}
\setCJKsansfont{HaranoAjiGothic}[
  Extension=.otf,
  UprightFont=*-Regular,
  BoldFont=*-Bold
]
\setCJKmonofont{HaranoAjiGothic}[
  Extension=.otf,
  UprightFont=*-Regular,
  BoldFont=*-Bold
]

\newcommand{\fixcolor}{\color{orange}}

\title{Text-Only vs. Image-Aware VLM Judges \\for Manga Translation Evaluation}

\author{
    \begin{tabular}{cc}
        \textbf{Subaru Kimura} & \textbf{Jun Suzuki} \\
        \multicolumn{2}{c}{Tohoku University} \\
        \multicolumn{2}{c}{\texttt{is-failab-research@grp.tohoku.ac.jp}}
    \end{tabular}
}

\begin{document}
\maketitle

\begin{abstract}
In manga translation, visual context outside speech-bubble text, including speaker identities, referents, and scene tone, is depicted in page images, making it natural to refer to images during translation evaluation.
However, most conventional manga translation evaluations rely solely on text-only automatic metrics, leaving the impact of image input insufficiently explored.
We propose a practical validation framework that employs the same vision-language model (VLM) as both a text-only judge (TOJ) and an image-aware judge (IAJ) to investigate evaluation divergences associated with image-aware evaluation procedures.
We use a Japanese-English manga dataset of official translations aligned at the page and speech-bubble levels.
First, adding page images to TOJ improved the primary VLM's detection of deliberately injected, visually dependent degradations.
Second, we analyzed evaluation divergences between TOJ and IAJ, finding different error counts and error-category distributions.
Human verification did not reveal a consistent preference between TOJ and IAJ.
\end{abstract}

\section{Introduction}
\label{sec:intro}

Machine translation research has advanced with the central goal of accurately converting text between languages at the sentence or document level.
In practice, however, what should be translated and which expressions are appropriate depend on the communicative situation, medium, and non-verbal information available to readers, which may not be captured by source and target text alone.

\begin{figure}[t]
  \centering
  \includegraphics[width=\linewidth]{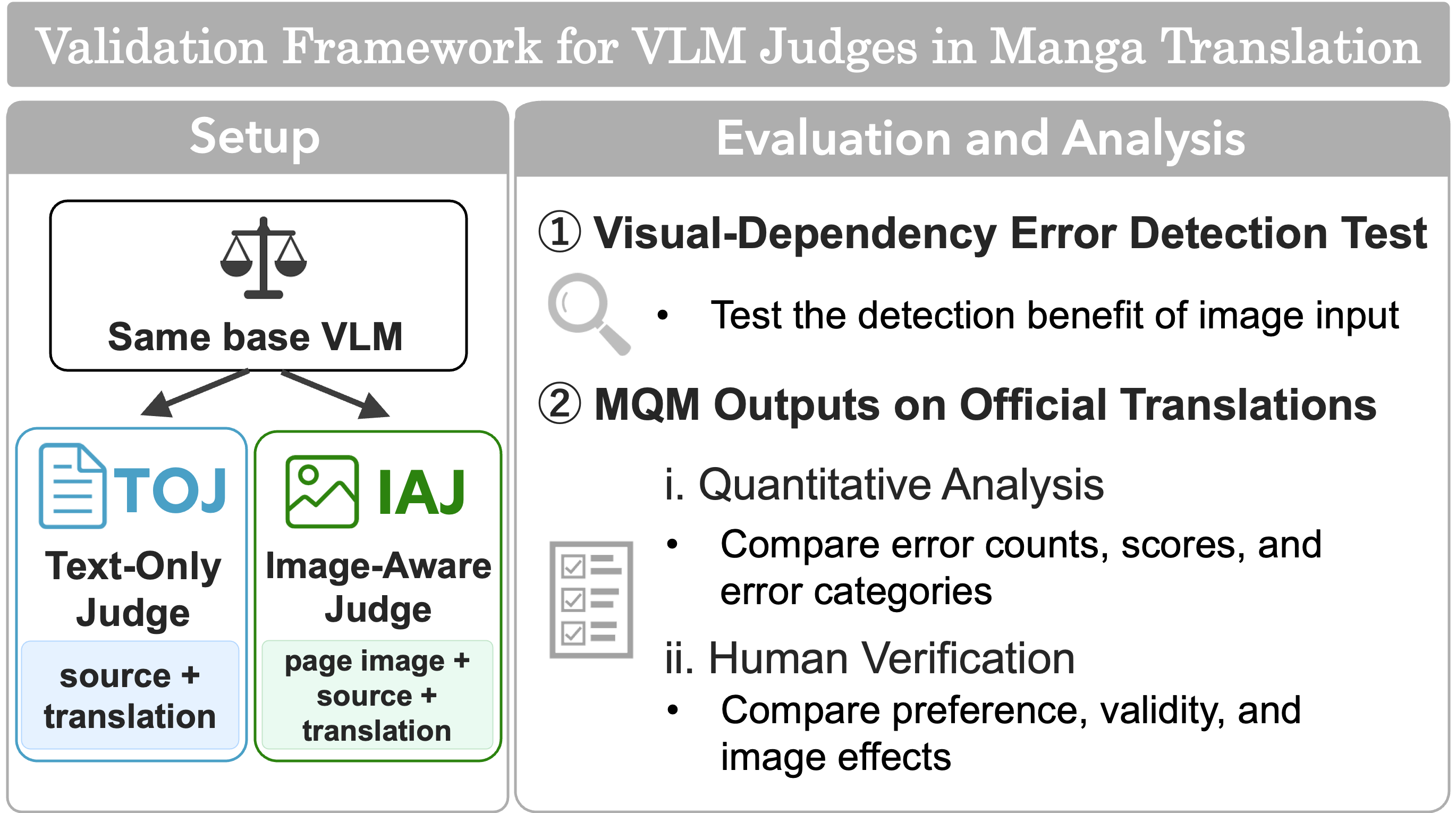}

  \caption{Overview of the proposed validation framework.}

  \label{fig:overview}
\end{figure}

Recent work has advanced both automatic translation evaluation \citep{lavie-etal-2025-wmt25-findings} and multimodal translation research incorporating visual context \citep{feng-etal-2025-mmnmt-survey}.
Manga translation is a prime example: page images depict not only dialogue text, but also speakers, referents, and characters' emotional states.

In manga translation quality evaluation (TQE), certain mistranslations, such as misidentifying speakers, pronouns, or object referents, or mismatching the tone of depicted scenes, can be difficult or impossible to identify reliably from source and target text alone and may require referencing page images.
In addition, in literary and manga translation, trade-offs exist between literal fidelity, naturalness, character persona, and visual context adaptation, such that a single reference translation cannot always serve as the unique ground truth.
Therefore, in manga TQE, including available page images in the input is natural to prevent information loss and evaluate appropriately.

Because VLMs can handle images and text simultaneously, an image-aware judge has the potential to detect visually dependent errors and incorporate context into evaluation that text alone cannot recover.
However, providing images does not necessarily make a VLM judge universally more accurate.
Errors in visual recognition or misinterpretations stemming from viewing images may also arise.
The goal of this paper is not to assert that IAJ is uniformly superior to TOJ, but to investigate how VLM judge behavior changes under image-aware evaluation procedures.

Robust validation would ideally involve large-scale human correlation studies that account for diversity across works, languages, and models.
However, human evaluation of manga translation demands domain expertise, and suitable evaluation data are scarce, making such validation difficult to conduct at scale.

To address this difficulty, we propose a practical validation framework that compares the same underlying VLM under two conditions: \textbf{Text-Only Judge} (TOJ), which evaluates source and target text, and \textbf{Image-Aware Judge} (IAJ), which incorporates page images into the evaluation and, in our default setting, additional visual-grounding instructions.
We use a Japanese-English manga dataset of official translations comprising 180 pages aligned at the speech-bubble level with page images.
Using this framework, we investigate three research questions: whether image input improves the detection of degraded translations requiring visual context, how automatic evaluations of TOJ and IAJ differ, and whether human verification reveals a consistent preference between their fine-grained outputs.

The overview of the proposed validation framework is shown in Figure~\ref{fig:overview}.

This paper makes four main contributions:
\begin{itemize}
\setlength{\itemsep}{0pt}
\item \textbf{Validation framework for VLM judges:}
We present a practical validation framework for manga translation evaluation using VLM judges that separately examines the effect of image input on detecting visually dependent errors and differences in evaluation characteristics between TOJ and IAJ on official translations.
\item \textbf{Manga translation dataset based on official translations:}
We transcribe Japanese source text and official English translations from manga pages publicly distributed by the rights holder, and construct an evaluation dataset of 180 pages aligned at the speech-bubble level with page images.
\item \textbf{Benefit from image input:}
In a degraded translation detection benchmark requiring visual context, simply adding page images to TOJ enables the primary VLM to detect more degraded translations.
\item \textbf{Analysis of evaluation differences and human verification between TOJ and IAJ:}
Comparing automatic evaluations of TOJ and IAJ on official translations, we show distinct trends in error counts, model-reported scores, and error-category distributions.
Furthermore, human verification by 6 annotators on 40 sampled pages with TOJ/IAJ output differences reveals no consistent preference for either TOJ or IAJ.
Post-task feedback suggests several directions for refining the human evaluation protocol, including prioritizing competing evaluation perspectives, defining boundary cases, and providing concrete examples of image effects.
\end{itemize}

\section{Related Work}
\label{sec:related}

\subsection{Machine Translation Evaluation and Large Language Model Judges}
In automatic machine translation (MT) evaluation, following lexical overlap metrics with reference translations like BLEU \citep{papineni-etal-2002-bleu} and chrF \citep{popovic-2015-chrf}, learned metrics such as COMET \citep{rei-etal-2020-comet} and MetricX \citep{juraska-etal-2023-metricx} have been developed.
Furthermore, large language model (LLM)-based metrics like G-Eval \citep{liu-etal-2023-g-eval} and GEMBA \citep{kocmi-federmann-2023-large} have enabled reference-free evaluation.

In human evaluation frameworks, Direct Assessment (DA) \citep{graham-etal-2013-continuous}, Error Span Annotation (ESA) \citep{kocmi-etal-2024-esa}, and Multidimensional Quality Metrics (MQM) \citep{lommel2014multidimensional,freitag-etal-2021-experts} have been widely adopted.
Recently, LLM-as-a-Judge using MQM-based evaluation has gained traction \citep{fernandes2023devil,kocmi-federmann-2023-gemba-mqm,agrawal-etal-2024-context-helpful}, and has shown efficacy in document-level, context-aware translation evaluation \citep{yan-etal-2025-nemo-wmt25,semenov-etal-2025-wmt25-terminology}.

Building on the established efficacy of LLM judges in MQM evaluation, this study investigates the efficacy of VLM judges in MQM evaluation.
We focus on differences between text-only and image-aware judges using the same underlying VLM.
This enables realistic, albeit bounded, validation of image input effects in automatic manga translation evaluation, which was previously difficult to verify.

\subsection{Multimodal Translation and Manga Translation}
As multimodal machine translation research progresses, translation methods leveraging visual context in manga translation have been proposed \citep{hinami2021towards,lippmann-etal-2025-context}.
In particular, Hinami et al. framed manga translation as a context-dependent multimodal task, showing that human evaluations viewing manga images rank translations differently than BLEU scores.
This provides an early example highlighting the limitations of evaluating manga translation using text-only metrics alone.
However, existing manga MT research focuses primarily on translation model development, leaving manga-specific TQE using page images, source text, and candidate translations underexplored.

Meanwhile, in the broader multimodal translation domain, studies leveraging VLM judges to analyze visual ambiguity in translation datasets have begun to emerge \citep{betala-etal-2025-picture}.
However, because it remains unclear whether models supplied with images actually use visual information, or what image input improves and changes, VLM judges themselves require rigorous validation.

Additionally, MangaVQA \citep{baek-etal-2026-mangavqa} demonstrates that recent VLMs possess foundational visual understanding capabilities for manga pages.
However, comprehending manga is necessary, but it is not equivalent to accurately evaluating translation quality.
Furthermore, even a single misrecognition can alter translation evaluation decisions.

Accordingly, we use the proposed validation framework to examine the impact of image input on VLM judges' detection of visually dependent errors and output characteristics in manga translation evaluation.

\section{Experimental Setup}
\label{sec:setup}

\subsection{Automatic Evaluation Design}
The evaluation unit is page-level. A single page is represented as a triplet $x = (I, s_{1:n}, t_{1:n})$, consisting of Japanese page image $I$, Japanese source segment sequence $s_{1:n}$, and English translation segment sequence $t_{1:n}$.
Here, $n$ denotes the number of speech bubbles (segments) on a page, varying by page.
TOJ takes $(s_{1:n}, t_{1:n})$ as input and uses a default MQM-style prompt.
Our default IAJ (IAJ-full) takes $(I, s_{1:n}, t_{1:n})$ as input and adds a visual-grounding instruction and a \texttt{visual\_caption} output requirement to the default prompt.
Note that the input image $I$ provided to IAJ is \textbf{only the Japanese page image}; the English target page image is not provided to the judge.
Both conditions employ MQM-style evaluation with our custom 10-category taxonomy, outputting an auxiliary model-reported score $S_p \in [0, 100]$ per page and a structured list of errors for each mistranslation location. Here, $S_p$ is the integer score output in the JSON field \texttt{quality\_score}. We define the corresponding model-reported score deficit as $D_p^{\mathrm{judge}} = 100 - S_p$. Separately, we compute fixed-weight penalties $D_p^{\mathrm{MQM}}$ from the three reported severity labels---\texttt{neutral}, \texttt{minor}, and \texttt{major}. Each error item consists of the corresponding segment index, error category, severity, and a natural language explanation.

Our MQM-style evaluation prompt adapts MQM-Chat \citep{li-etal-2025-mqm} (which defines 7 error types for dialogue translation, including ambiguity, slang/loanwords, and dialogue consistency), redefining and expanding categories to match manga translation characteristics (segment-level bubbles, personal reference ambiguity, tone/emotion, and visual scene alignment) into our custom 10-category taxonomy (full prompt in Appendix~\ref{sec:appendix-prompts}).

\subsection{Dataset}
As the evaluation target, we use Volume 1 of \emph{Give My Regards to Black Jack}, by Shuho Sato.
In this study, we transcribed Japanese source text and English translation text from publicly available manga images, establishing 180 pages with unique 1:1 segment alignment at the bubble level verified via optical character recognition (OCR) and human inspection (dataset statistics in Appendix~\ref{sec:appendix-dataset-stats}).

The evaluation data are publicly available at \url{https://huggingface.co/datasets/joddy/gmr2bj_manga_mt_ja_en} for reproducibility.
Details of released dataset contents are provided in Appendix~\ref{sec:data-release}.

\subsection{Models}
The primary evaluation model is \texttt{gemini-\allowbreak{}3-\allowbreak{}flash-\allowbreak{}preview} \citep{gemini3flash}. This selection is based on pilot study findings showing that this model balances manga page image comprehension and structured output stability effectively. Furthermore, MangaVQA reports strong performance of Gemini-family models on manga page understanding \citep{baek-etal-2026-mangavqa}, which motivated our use of a Gemini-family VLM in this study.
We employ the same model under both TOJ and IAJ conditions to analyze differences between text-only and image-aware evaluation.

In this paper, we primarily report results for the main evaluation model.
To verify that gains from image-aware procedures are not unique to the primary model, we also perform automatic MQM evaluation with \texttt{gpt-5.4-mini} \citep{gpt54mini} and \texttt{qwen3-vl-32b-instruct} \citep{bai2025qwen3vl} as a robustness check.
Experiment 1 evaluates 75 degraded translations, and Experiment 2 evaluates 180 official translation pages.

\section{Experiment 1: Degraded Translation Detection}
\label{sec:exp1}

\subsection{Creation of Degraded Translations}
To intentionally generate ``translation errors that cannot be decisively identified without viewing page images,'' we designed four types of degradation operations shown in Table~\ref{tab:degradation-types}.
These four degradation types are benchmark-construction categories and are distinct from the ten MQM error categories used by the judges.
Using \texttt{gemini-2.5-pro} \citep{comanici2025gemini25} and \texttt{gemini-3-flash-preview}, we generated degraded translation candidates contradicting image content across these four types, selecting a total of 75 benchmark instances through human vetting.
Across the entire process, 75 final instances were adopted from over 900 generated candidates, yielding an overall adoption rate of approximately 8\%.
We note that the number and variety of highly visually dependent degradations that can be constructed from a single 180-page volume are inherently limited.
Prompts used for candidate creation are detailed in Appendix~\ref{sec:appendix-prompts}.

\begin{table}[t]
\centering
\footnotesize
\setlength{\tabcolsep}{8pt}
\renewcommand{\arraystretch}{1.2}
\begin{tabular}{@{}p{0.38\columnwidth}p{0.54\columnwidth}@{}}
\toprule
\textbf{Degradation Type} & \textbf{Description} \\
\midrule
\parbox[t]{29mm}{\raggedright\ttfamily speaker\_or\_\\addressee} & Confuses speaker or addressee, shifting subjects or honorifics \\
\parbox[t]{29mm}{\raggedright\ttfamily referent\_or\_\\object} & Translates referenced objects/entities in image as something else \\
\parbox[t]{29mm}{\raggedright\ttfamily tone\_emotion\_or\_\\style} & Translates with tone or style contradicting facial expression or scene \\
\parbox[t]{29mm}{\raggedright\ttfamily action\_or\_\\state} & Translates with actions or physical states contradicting depicted artwork \\
\bottomrule
\end{tabular}
\caption{Degradation operation types used in Experiment 1 to create translations contradicting manga page image content.}
\label{tab:degradation-types}
\end{table}

\subsection{Metrics}
In Experiment 1, we post-hoc evaluate whether error explanations output by TOJ/IAJ as MQM-style judges using our custom 10-category taxonomy semantically capture the pre-designed degradation intent.

For each degraded translation, \texttt{gpt-5.6-luna} \citep{openai2026gpt56} classifies the relationship between the judge's error explanation and the degradation intent into \{\texttt{same}, \texttt{related}, \texttt{different}\}, counting \texttt{same} as a semantic match. We term this proportion the \textbf{semantic match rate} (prompts in Appendix~\ref{sec:appendix-prompts}).
This is an instance-level detection rate: the proportion of the 75 degraded translations for which at least one reported error is classified as \texttt{same}. It is not conventional overall accuracy, error-level recall, or MQM accuracy.
This metric does not assess the precision of the reported errors.
Reported errors that do not match the injected degradation may still identify genuine translation issues and therefore cannot automatically be treated as false positives.
Without independent annotations of these additional errors, we cannot reliably compute precision or an F1 score.
We also report the mean number of error flags per degraded translation, computed from the 75 fixed MQM outputs independently of the five semantic match runs.
During exploratory stages, we experimented with rule-based metrics, but selected semantic match as an LLM-based evaluation method capable of accounting for paraphrastic variation.
We formulate semantic matching as a natural language inference (NLI)-like classification task that can accommodate paraphrased explanations. Its validity remains limited by the lack of independently annotated gold labels.

\subsection{Compared Evaluation Conditions}
The prompt used for our default IAJ (IAJ-full) adds a visual-grounding instruction and a \texttt{visual\_caption} requirement to TOJ's base prompt.
The difference between TOJ and IAJ-full encompasses not only image presence, but also the combined effect of input modality and visual grounding procedures.
To disentangle this confounding factor, Experiment 1 introduces \textbf{IAJ-minimal} (adding only page images to TOJ) and \textbf{IAJ-blank} (replacing IAJ-full images with a blank image).

\begin{table*}[!t]
\centering
\small
\input{tables/en/experiment1_factorial_ablation}
\caption{Experiment 1 results. Semantic match rates are summarized by their mean and range across five runs. Mean flags denotes the average number of reported errors per instance.}
\label{tab:exp1-main}
\end{table*}

\subsection{Results}
Because semantic match judgments can vary across runs, we repeated semantic match classification five times using the same fixed MQM outputs, without re-running the underlying VLM judges.
We report the mean and range in Table~\ref{tab:exp1-main}.
IAJ-minimal achieved a mean rate of 81.6\%, and IAJ-full achieved 80.5\%; both had higher semantic match rates than TOJ's 57.9\%.
Conversely, IAJ-blank achieved a mean rate of 52.0\%, dropping below both IAJ-full and TOJ.
The pure contrast of image presence (TOJ vs. IAJ-minimal) yielded a 5-run mean difference of +23.7 percentage points. Comparing identical degraded translations between TOJ and IAJ-minimal and testing aggregated differences grouped by source page revealed that IAJ-minimal's gain was statistically significant across all 5 runs. Significance test details are shown in Appendix~\ref{sec:appendix-exp1-significance}.
This ablation result shows, on this constructed benchmark, that providing page images increases the semantic match rate compared to TOJ, whereas substituting actual images with blank images degrades the semantic match rate.
This provides evidence that page images help detect degraded translations requiring visual context on this constructed benchmark.

The slightly higher semantic match rate of IAJ-minimal over IAJ-full does not necessarily imply that IAJ-minimal is a more accurate judge.
Semantic match measures whether an error explanation matching the degradation intent is contained in the MQM output, not overall MQM precision.
IAJ-minimal produced more error flags overall than IAJ-full, which may have driven up the semantic match rate.
Furthermore, we note that the additional prompt in IAJ-full was added to the base prompt because preliminary qualitative checks showed it aided MQM precision prior to scaling experiments.

\begin{table*}[!t]
\centering
\footnotesize
\setlength{\tabcolsep}{4pt}
\begin{tabular}{lrrrrr}
\toprule
 & \multicolumn{2}{c}{\textbf{Semantic match rate}} & & \multicolumn{2}{c}{\textbf{Mean flags}} \\
\cmidrule(lr){2-3}\cmidrule(lr){5-6}
\textbf{MQM Judge} & \textbf{TOJ} & \textbf{IAJ-full} & \textbf{Difference} & \textbf{TOJ} & \textbf{IAJ-full} \\
\midrule
\texttt{gemini-3-flash-preview} & 57.9\% (56.0--58.7) & 80.5\% (78.7--82.7) & +22.7 pp & 1.11 & 1.20 \\
\texttt{gpt-5.4-mini} & 24.8\% (24.0--25.3) & 30.7\% (30.7--30.7) & +5.9 pp & 1.05 & 0.89 \\
\texttt{qwen3-vl-32b-instruct} & 45.6\% (44.0--48.0) & 59.2\% (57.3--61.3) & +13.6 pp & 1.43 & 1.53 \\
\bottomrule
\end{tabular}
\caption{Experiment 1 cross-model comparison. Semantic match rates are five-run means with ranges in parentheses. Differences are IAJ-full minus TOJ, computed before rounding. Mean flags is defined as in Table~\ref{tab:exp1-main}.}
\label{tab:exp1-robustness}
\end{table*}

We illustrate one representative degraded translation contradicting manga image content that was detected by IAJ-full but missed by TOJ. This example concerns a visually dependent degradation involving a referent. Additional examples of visually dependent degradations are provided in Figure~\ref{fig:exp1-pages} in Appendix~\ref{sec:appendix-exp1}.

\begin{figure*}[!t]
\centering
\begin{minipage}[t]{0.3\textwidth}
  \vspace{0pt}
  \centering
  \includegraphics[width=\linewidth,height=5.0cm,keepaspectratio]{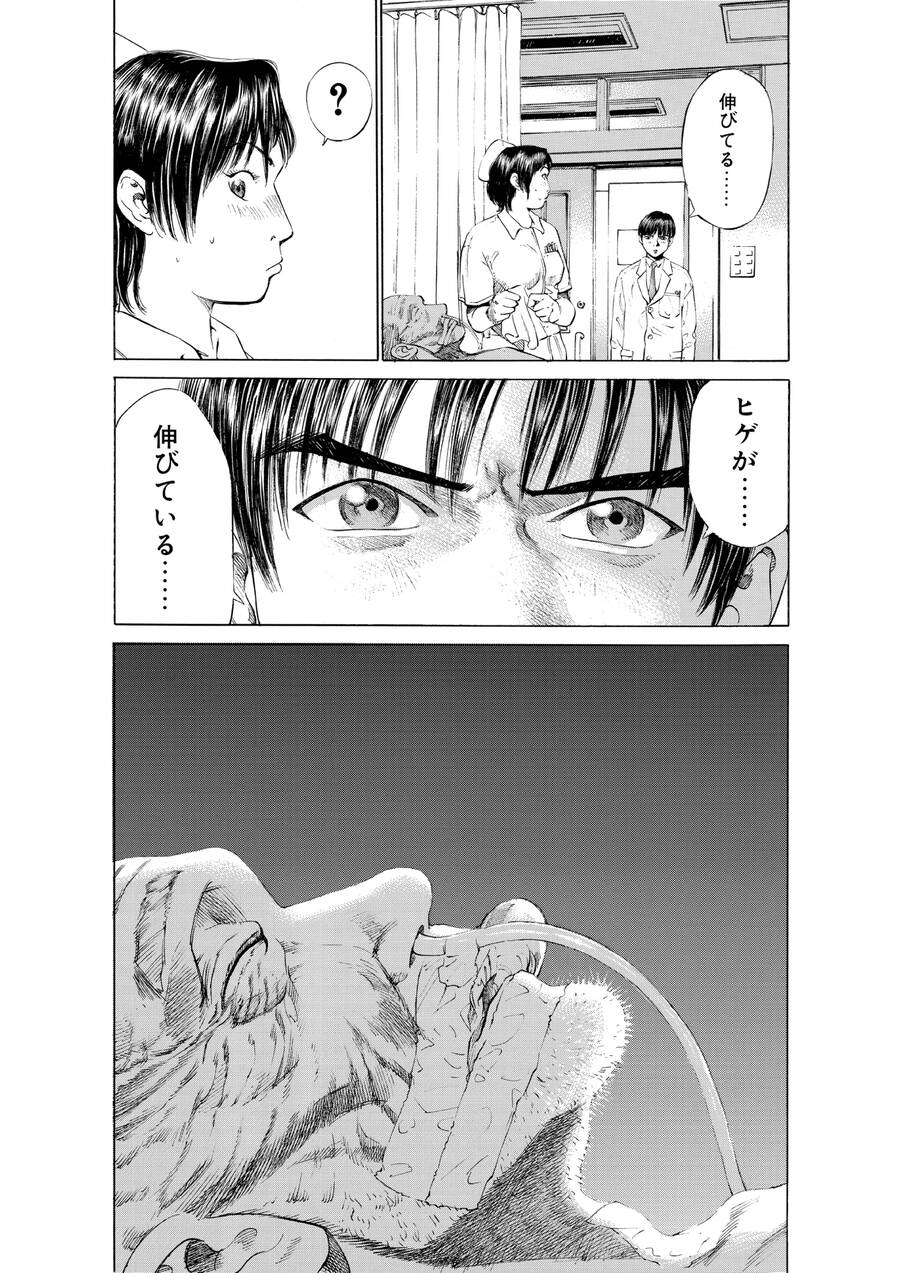}
  \caption*{\footnotesize bj\_vol1\_113 (JA)}
\end{minipage}
\hfill
\begin{minipage}[t]{0.3\textwidth}
  \vspace{0pt}
  \centering
  \includegraphics[width=\linewidth,height=5.0cm,keepaspectratio]{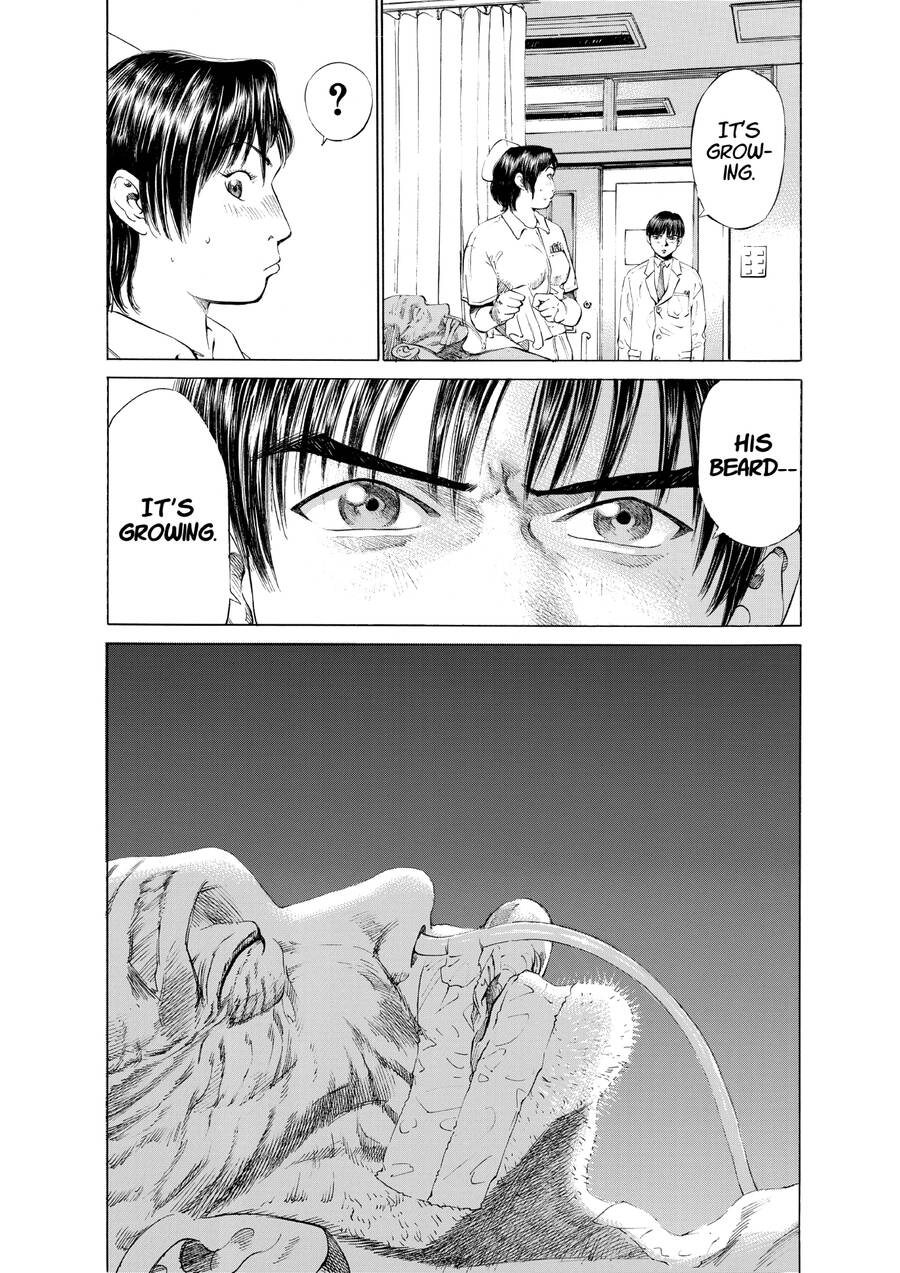}
  \caption*{\footnotesize bj\_vol1\_113 (EN)}
\end{minipage}
\hfill
\begin{minipage}[t]{0.38\textwidth}
  \vspace{0pt}

  \footnotesize
  \textbf{Degraded segment}\\
  type: \texttt{referent\_or\_object}\\
  JP ``ヒゲが……'' \\
  \texttt{\hspace*{1em}} EN (original) ``HIS BEARD--''\\
  \texttt{\hspace*{1em}} EN (degraded) ``MY BEARD--''\\

  \textbf{Degradation intent}\\
  ``The possessive `HIS' was changed to `MY', shifting the referent of the beard from the patient to the speaker (the doctor).''\\

  \textbf{IAJ-full error explanation}\\
  ``The doctor is observing the patient's beard, not his own. The use of the first-person possessive `MY' is a reference error that misidentifies whose beard is being discussed.''
\end{minipage}
\caption{Representative example (\texttt{bj\_vol1\_113}) of degraded translation contradicting manga image content missed by TOJ but detected by IAJ-full. Left: JA/EN page images; Right: degraded segment, degradation intent, and IAJ-full error explanation. The EN page image is shown for illustration only and was not provided to IAJ. The patient's beard and the doctor's face clarify the referent error changing ``HIS'' to ``MY''. \textit{Give My Regards to Black Jack / SHUHO SATO}}
\label{fig:exp1-pages-main}
\end{figure*}

\subsection{Robustness Across Other VLMs}
To verify whether divergences manifest across models, we evaluated TOJ and IAJ-full conditions using \texttt{gpt-\allowbreak{}5.4-\allowbreak{}mini} and \texttt{qwen3-\allowbreak{}vl-\allowbreak{}32b-\allowbreak{}instruct} (Table~\ref{tab:exp1-robustness}).

In both models, IAJ-full achieved a higher semantic match rate than TOJ, though the rates varied considerably by model.
These results show that the directional divergence between TOJ and IAJ-full is not unique to the primary model, while indicating that absolute semantic match rates and gain magnitudes depend on the underlying VLM.

\section{Experiment 2: TOJ/IAJ Evaluation Characteristics and Human Verification}
\label{sec:exp2}
In Experiment 2, we deliberately evaluate original official translations using automatic MQM-style evaluation to quantitatively and qualitatively analyze output differences between TOJ and IAJ-full.
Hereafter, references to IAJ imply IAJ-full unless noted otherwise.
When treating official translations as a provisional reference baseline, errors reported by judges can be interpreted as provisional false positives relative to the official translation or as overflagging relative to that baseline.
However, official translations are not absolute ground truth; they may contain actual translation errors or room for improvement, and alternative valid translations exist.
Furthermore, discrepancies between choices made in official translations and evaluation aspects prioritized by TOJ/IAJ can cause judges to flag errors in official translations.
Thus, evaluating official translations allows us to analyze the evaluation characteristics of TOJ and IAJ.

\subsection{Quantitative Analysis Design}
By analyzing differences in MQM-style outputs between TOJ and IAJ-full, we quantitatively elucidate differences in their evaluation characteristics.
Errors in MQM-style outputs are categorized into TOJ-only, IAJ-only, and common, analyzing output divergences.
We match TOJ and IAJ flags at the segment level using a rule-based procedure, classifying corresponding flags as common and non-corresponding flags as TOJ-only or IAJ-only.
This heuristic matching combines explanation token set Jaccard similarity and string similarity with \texttt{error\_category} and \texttt{severity} alignment.
The objective is to mechanically link obviously identical flags to aggregate output differences quantitatively.
Therefore, this classification does not guarantee strict semantic identity, and results should be interpreted as diagnostic statistics describing TOJ and IAJ output divergences.
Matching methodology details are provided in Section~\ref{sec:appendix-base-rate} of the Appendix.

\subsection{Qualitative Analysis Design}
To qualitatively analyze TOJ and IAJ-full evaluations, we conduct human verification involving 6 Japanese annotators from the same laboratory (non-authors).
Annotators were native Japanese speakers familiar with reading manga and possessing backgrounds in language processing research.
All annotators were master's or doctoral students at Tohoku University and had no prior experience in professional translation or MQM annotation.
Standardized English proficiency test scores were not collected.
While using professional Japanese-to-English manga translators would be ideal, recruiting and executing verification with multiple professional translators presents substantial practical hurdles.

For human verification targets, from the candidate set of 116 pages among 180 pages having at least one unique flag in TOJ or IAJ, we selected 40 pages via simple random sampling.
Sampling from pages with output differences allows for a clearer analysis of evaluation characteristic divergences between TOJ and IAJ.
The selected 40 pages comprised 12 pages with unique flags in TOJ only, 4 pages in IAJ only, and 24 pages with unique flags in both.
For these 40 pages, evaluators compared TOJ and IAJ flags for the same page.

The main judgment asked evaluators which judge's flags were superior, with three response options: \texttt{prefer\_IAJ} / \texttt{prefer\_TOJ} / \texttt{tie}.
Additionally, evaluators reported confidence, page-level output validity ratings for TOJ/IAJ, image effect on IAJ flags, and usefulness of image information in translation evaluation. Each validity rating covered the complete flag set produced by one judge for one page; thus, the denominator was $40$ pages $\times$ $6$ annotators $=240$ ratings per judge. Evaluators received descriptions for each option but were not instructed on a rigid priority hierarchy when source fidelity, naturalness, and visual consistency conflicted. When 4 or more of 6 evaluators selected the same main judgment, it was classified as strict majority; otherwise, it was classified as \texttt{no\_majority}.
For operational protocol details, annotation user interface (UI), and instructions, see Appendix~\ref{sec:appendix-annotators}.

\subsection{Quantitative Analysis of MQM-style Evaluation}
\subsubsection{Overview of Automatic Evaluation across 180 Pages}
Table~\ref{tab:exp2-base-rate-main} shows automatic evaluation results across all 180 official translation pages for TOJ and IAJ-full, including error-flag counts and model-reported score statistics.
For gemini-3-flash-preview, IAJ-full yielded fewer mean error flags per page than TOJ ($1.77 \to 1.46$), and a lower mean model-reported score deficit ($100-S_p$) ($10.24 \to 8.08$).
Each reported error is assigned one of three severities: \texttt{major}, \texttt{minor}, or \texttt{neutral}. Prompts define \texttt{major} as errors substantially distorting meaning, \texttt{minor} as errors with limited impact, and \texttt{neutral} as reported issues with constrained effect. In this paper, all three severities are included in error flag counts.
Major flags were fewer than minor flags, with major-free pages accounting for 147/180 (81.7\%) in TOJ and 158/180 (87.8\%) in IAJ-full. Conversely, minor errors occurred at 1--2 items per page across most pages for both judges.
While this does not directly show that IAJ-full produced more accurate evaluations, it shows that image-aware procedures alter judge output volume and reported scores.

As a fixed-weight scoring analysis based on MQM severity labels, we assigned weights of 0, 1, and 5 to neutral, minor, and major, respectively, and calculated penalty points per 100 words of the official English translation. For the primary evaluation model, TOJ yielded 5.80 penalty points, compared with 4.57 for IAJ-full; thus, the tendency for IAJ-full to incur fewer penalties was reproduced in the fixed-weight MQM aggregation. This result is consistent with the comparison based on the judge-reported \texttt{quality\_score}; the page-level Spearman correlations between the two measures were 0.762 for TOJ and 0.717 for IAJ-full. The auxiliary models showed differences in the same direction, while fixed-weight scoring and sensitivity analyses are provided in Appendix~\ref{sec:appendix-base-rate}.

Thus, automatic comparison across 180 pages showed that IAJ-full produced fewer flags and smaller auxiliary score deficits than TOJ.
Auxiliary models exhibited a similar directional trend: IAJ-full produced fewer total error flags and smaller auxiliary mean score deficits than TOJ.
Auxiliary model results for Experiment 2 are detailed in Appendix~\ref{sec:appendix-exp2-robustness}.

\begin{table}[!htbp]
\centering
\small
\setlength{\tabcolsep}{4pt}
\begin{tabular}{@{}lrr@{}}
\toprule
\textbf{Metric} & \textbf{TOJ} & \textbf{IAJ-full} \\
\midrule
\multicolumn{3}{l}{\emph{All 180 Pages}} \\
Total error flags & 318 & 262 \\
Mean / page & 1.77 & 1.46 \\
Major error flags & 36 & 24 \\
Mean major / page & 0.20 & 0.13 \\
Minor error flags & 268 & 233 \\
Mean minor / page & 1.49 & 1.29 \\
Neutral error flags & 14 & 5 \\
Mean neutral / page & 0.08 & 0.03 \\
Mean quality\_score & 89.8 & 91.9 \\
Mean score deficit ($100-S_p$) & 10.24 & 8.08 \\
\midrule
\multicolumn{3}{l}{\emph{Page-Level Error Distribution}} \\
0-error pages & 20 (11.1\%) & 31 (17.2\%) \\
1+ error pages & 160 (88.9\%) & 149 (82.8\%) \\
0-major pages & 147 (81.7\%) & 158 (87.8\%) \\
0-error in both judges & \multicolumn{2}{c}{19 (10.6\%)} \\
\bottomrule
\end{tabular}
\caption{Automatic evaluation results reported by judges on official translations across 180 pages (based on \texttt{gemini-3-flash-preview}). The table includes total error flags, severity counts, mean model-reported \texttt{quality\_score}, and mean score deficit ($100-S_p$). Total error flags include \texttt{major}, \texttt{minor}, and \texttt{neutral}.}
\label{tab:exp2-base-rate-main}
\end{table}

\begin{table*}[!t]
\centering
\small
\begin{tabular}{p{0.38\textwidth}rrr}
\toprule
\textbf{Category} & \textbf{TOJ-only} & \textbf{IAJ-only} & \textbf{TOJ-only Share} \\
\midrule
Overall & 152 & 96 & 61.3\% \\
\midrule
meaning\_accuracy & 42 & 32 & 56.8\% \\
naturalness\_and\_register & 30 & 19 & 61.2\% \\
omission\_or\_addition & 31 & 17 & 64.6\% \\
tone\_and\_emotion & 19 & 10 & 65.5\% \\
terminology\_and\_named\_entities & 20 & 12 & 62.5\% \\
reference\_and\_disambiguation & 4 & 4 & 50.0\% \\
action\_and\_state & 1 & 1 & 50.0\% \\
speaker\_and\_dialogue\_consistency & 0 & 0 & -- \\
colloquial\_and\_cultural\_expression & 3 & 0 & 100.0\% \\
other & 2 & 1 & 66.7\% \\
\bottomrule
\end{tabular}
\caption{Breakdown of TOJ/IAJ unique error flags by MQM category across 180 pages.}
\label{tab:exp2-disagreement-by-category}
\end{table*}

\subsubsection{Structural Analysis of Judge Output Divergence}
To examine TOJ and IAJ-full output differences in detail, we categorized errors across all 180 pages into TOJ-only, IAJ-only, and common.
Overall, TOJ-only = 152, IAJ-only = 96, and common = 166, showing TOJ produced more unique flags. We define TOJ-only share as $\text{TOJ-only}/(\text{TOJ-only}+\text{IAJ-only})$. In \texttt{tone\_\allowbreak{}and\_\allowbreak{}emotion} (65.5\%), \texttt{omission\_\allowbreak{}or\_\allowbreak{}addition} (64.6\%), \texttt{terminology\_\allowbreak{}and\_\allowbreak{}named\_\allowbreak{}entities} (62.5\%), and \texttt{naturalness\_\allowbreak{}and\_\allowbreak{}register} (61.2\%), TOJ-only flags were more frequent than IAJ-only flags, suggesting that TOJ often flags fine-grained differences in naturalness, style, and fidelity.
In contrast, \texttt{reference\_and\_disambiguation} and \texttt{action\_and\_state} showed nearly equal TOJ-only and IAJ-only shares, indicating IAJ-full also produced unique flags regarding visual reference resolution and action understanding.
Category breakdowns for TOJ-only / IAJ-only are presented in Table~\ref{tab:exp2-disagreement-by-category}.

Furthermore, comparing 54 segment locations where both judges flagged the same page segment but could not be matched as common, 37 instances (68.5\%) differed in \texttt{error\_category}.
This indicates that even when targeting the same segment, TOJ and IAJ frequently conceptualize issues from different perspectives.

\subsubsection{Retention of TOJ Flags in IAJ-full}
We define coverage as the proportion of error candidates flagged by TOJ that are retained as identical flags in IAJ-full.
Across 160 pages with 1+ TOJ flag (totaling 318 TOJ flags), 166 flags (52.2\%) were retained in IAJ-full. On 45 pages (28.1\%), IAJ-full retained zero TOJ flags.
This indicates that while IAJ-full incorporates flags overlapping with TOJ, it also does not retain some flags reported by TOJ.
Detailed coverage distribution statistics are provided in Table~\ref{tab:coverage-distribution} in Appendix~\ref{sec:appendix-coverage}.

\begin{table}[!htbp]
\centering
\footnotesize
\begin{tabular}{@{}lr@{}}
\toprule
\textbf{Metric} & \textbf{Value} \\
\midrule
Annotators / Verified Pages & 6 / 40 \\
\midrule
\multicolumn{2}{l}{\textit{Individual votes ($n=240$)}} \\
Prefer IAJ / TOJ / Tie & 73 / 87 / 80 \\
\quad Percentage & 30.4 / 36.3 / 33.3\% \\
Medium or high confidence & 228 (95.0\%) \\
\midrule
Mean Pairwise Agreement & 0.512 \\
Fleiss' $\kappa$ & 0.266 \\
Strict Majority - unanimous & 5 / 40 \\
Strict Majority - non-unanimous & 21 / 40 \\
Any majority & 26 / 40 (65.0\%) \\
\midrule
\texttt{prefer\_IAJ} Consensus & 7 \\
\texttt{prefer\_TOJ} Consensus & 12 \\
\texttt{tie} Consensus & 7 \\
\texttt{no\_majority} & 14 / 40 (35.0\%) \\
\bottomrule
\end{tabular}
\caption{Experiment 2: Main judgment agreement and page-level consensus among 6 annotators over 40 TOJ/IAJ-full MQM outputs.}
\label{tab:exp2-main}
\end{table}

\begin{table*}[!t]
\centering
\footnotesize
\setlength{\tabcolsep}{4pt}
\begin{minipage}[t]{0.49\textwidth}
\vspace{0pt}
\centering
\begin{tabular}{@{}lrr@{}}
\toprule
\textbf{Output validity ($n=240$ each)} & \textbf{TOJ} & \textbf{IAJ} \\
 & \textbf{Count (\%)} & \textbf{Count (\%)} \\
\midrule
Valid & 168 (70.0) & 172 (71.7) \\
Partially valid & 46 (19.2) & 28 (11.7) \\
Invalid & 15 (6.3) & 12 (5.0) \\
Unclear & 11 (4.6) & 28 (11.7) \\
Valid or partially valid & 214 (89.2) & 200 (83.3) \\
\midrule
\textbf{Tie reason ($n=80$)} & \multicolumn{2}{r}{\textbf{Count (\%)}} \\
\midrule
Both valid & \multicolumn{2}{r}{58 (72.5)} \\
Both partially valid & \multicolumn{2}{r}{10 (12.5)} \\
Difference too small & \multicolumn{2}{r}{10 (12.5)} \\
Both invalid & \multicolumn{2}{r}{1 (1.3)} \\
Other & \multicolumn{2}{r}{1 (1.3)} \\
\bottomrule
\end{tabular}
\end{minipage}\hfill
\begin{minipage}[t]{0.49\textwidth}
\vspace{0pt}
\centering
\begin{tabular}{@{}lr@{}}
\toprule
\textbf{Image usefulness ($n=240$)} & \textbf{Count (\%)} \\
\midrule
Helpful or essential & 124 (51.7) \\
Not needed & 109 (45.4) \\
Unclear & 7 (2.9) \\
\midrule
\textbf{Image effect ($n=240$)} & \textbf{Count (\%)} \\
\midrule
No observable effect & 137 (57.1) \\
Any positive effect & 55 (22.9) \\
Any negative effect & 26 (10.8) \\
Unclear & 24 (10.0) \\
\bottomrule
\end{tabular}
\end{minipage}
\caption{Experiment 2 human ratings. Left: output validity and reasons for ties. Right: perceived image usefulness and observed effects on IAJ outputs. Values are counts with percentages in parentheses. IAJ denotes IAJ-full. Positive effects include valid detection, overflagging prevention, or rationale strengthening; negative effects include visual misrecognition/overinterpretation or missed textual issues. Effects are counted per response, with two responses reporting both positive and negative effects.}
\label{tab:exp2-human-details}
\end{table*}

\subsection{Qualitative Analysis via Human Verification}
\subsubsection{Main Judgment and Agreement among 6 Annotators}
Table~\ref{tab:exp2-main} summarizes the 240 judgments from 6 annotators over 40 pages. Preferences were dispersed, and agreement was limited despite predominantly medium or high confidence. Among the 14 pages without a majority, 12 had a 3--2--1 split and two a 3--3 split; most disagreements were therefore not clean binary divisions.

Annotators also showed distinct tendencies: one chose \texttt{tie} on 27 pages, another preferred TOJ on 24, and another preferred IAJ on 17. Low agreement thus cannot be explained merely by hesitation. Possible factors include small differences between outputs, differing criteria or tie thresholds, and characteristics of the sampled pages, but these results do not isolate their contributions.

\subsubsection{Page-Level Output Validity and Reasons for Tie}
Table~\ref{tab:exp2-human-details} summarizes ratings of each judge's complete page-level flag set, not individual errors. Both judges were often rated valid, with similar valid rates, whereas IAJ received more unclear ratings. The dominant tie reason was that both sets were valid, rather than both invalid. Rating definitions are provided in Appendix~\ref{sec:appendix-annotators}.

\subsubsection{Usefulness of Image Information and Effect on IAJ}
Perceived image usefulness and observed effects on IAJ flags address different questions (Table~\ref{tab:exp2-human-details}). Judgments of potential usefulness were nearly evenly divided, whereas most responses reported no observable effect on IAJ flags. Positive and negative effects were both reported, indicating instance-dependent benefits and occasional adverse effects rather than consistent improvement. Image effects allowed multiple selections; full option definitions appear in Appendix~\ref{sec:appendix-annotators}.

\subsubsection{Post-Task Feedback}
As post-task feedback upon completing annotations, evaluators answered 5-point scale items regarding overall task difficulty, rubric sufficiency, priority clarity when criteria conflict (source fidelity vs. target naturalness vs. visual consistency), main label clarity, and usefulness of concrete examples. Additionally, evaluators picked the single annotation field that caused the most hesitation across 40 pages, and selected multiple-choice options for explanations or examples that would aid judgment in future iterations (e.g., priority rules, boundary case examples, image effect examples, severity rules, concrete label examples).
Across 6 annotators' post-task feedback, median task difficulty was 4.0/5, while rubric sufficiency and main label clarity medians were both 4.0. In contrast, priority clarity when criteria conflict received a median of only 2.5. Usefulness of concrete label examples or boundary cases had a median of 4.0, and 4 of 6 annotators picked image effect classification as the hardest field. Additional guidance covering priority rules, boundary examples, image effect examples, severity examples, and label examples was requested by 3 annotators each. Thus, future evaluation designs should consider providing explicit priority rules and concrete benchmark examples when evaluation perspectives conflict. However, causal improvements in agreement or accuracy from adding guidelines remain unverified.

\section{Discussion}
\label{sec:discussion}
We briefly summarize key insights derived from experimental results.

\paragraph{Does Adding Page Images Aid Visually Dependent Error Detection?}
In Experiment 1, for 75 degraded translations designed to be difficult to judge without images, IAJ-minimal (adding only page images to TOJ) achieved a higher mean semantic match rate than TOJ, with the difference being statistically significant across all 5 runs.
These results provide evidence that, on this constructed benchmark, inputting page images aids the detection of visually dependent degradations.

\paragraph{How Does Image-Aware Procedure Alter Judge Evaluation Characteristics?}
In Experiment 2, IAJ-full produced fewer flags and smaller auxiliary score deficits derived from model-reported scores than TOJ, exhibiting divergences in unique flag counts and error category assignments for identical segments. This indicates that IAJ-full does not merely add flags: some flags reported by TOJ are not retained, and errors are reported from distinct angles.

\paragraph{Can Human Verification Stably Arbitrate TOJ vs. IAJ Differences?}
Human verification showed no consistent preference trend favoring either TOJ or IAJ. Page-level output validity ratings for both conditions did not differ markedly, and \texttt{both\_valid} was the dominant reason for \texttt{tie} judgments. This pattern is consistent with the difficulty of ranking two flag sets both judged valid overall, although the present results do not isolate the cause of disagreement.
Evaluator perceptions of potential image usefulness split evenly, while observed image effects on actual IAJ flags showed no effect in 57.1\%, positive effects in 22.9\%, and negative effects in 10.8\%.
Thus, human verification found no uniform preference for IAJ-full, indicating that the image-aware evaluation procedure does not yield uniformly preferred evaluations across all pages.
Post-task feedback showed that annotators sometimes had difficulty choosing between judge outputs when different evaluation criteria favored different outputs.
The limited agreement may partly reflect differences in how annotators prioritize source fidelity, target-language naturalness, and visual consistency when comparing judge outputs.
Our protocol did not prescribe a strict priority among these criteria, leaving room for differing judgments.
However, the present results cannot determine whether disagreement primarily reflects small differences between judge outputs, differing priorities among evaluation criteria, or characteristics of the manga title and sampled pages.

\paragraph{What Is Required for Practical VLM Judge Evaluation?}
Future work should calibrate evaluators after sharing priority rules for conflicting evaluation perspectives, common criteria, boundary cases, and concrete examples of image effects. However, whether adding such explanations or calibration improves agreement or judgment stability remains unverified. Practical use also requires additional validation across multiple works, other language pairs, and models.

\section{Conclusion}
\label{sec:conclusion}

This study presents a validation framework for manga translation VLM judges that separately evaluates detection gains for visually dependent errors and characteristic output changes induced by image-aware procedures.
To accomplish this, we built an open evaluation dataset of 180 pages from transcribed official Japanese and English manga, aligned at the speech-bubble level with page images.
In Experiment 1, we found that adding page images aids in detecting visually dependent errors on this constructed benchmark, providing evidence for a baseline benefit of utilizing VLM judges.
In Experiment 2, while IAJ-full exhibited different error volume and evaluation perspectives from TOJ, human preference judgments across 6 annotators showed no consistent superiority.
These findings suggest that IAJ merits further investigation for manga translation evaluation, but do not support replacing TOJ with IAJ universally.
Through our proposed validation framework, we present a practical validation approach for settings where large-scale human correlation studies are difficult.
Our framework provides a starting point for extending VLM judge evaluation to other visually dependent translation domains, such as subtitles, advertisements, and instruction manuals.

\clearpage
\section*{Limitations}
\begin{itemize}
\item
The evaluation target is restricted to a single manga work and language pair (ja$\to$en). Generalizability to other works, genres, or language pairs remains unverified; we do not claim identical effect sizes or output characteristics hold across manga broadly. Copyright constraints surrounding redistributable official parallel manga data exacerbate this limitation.
\item
Degraded translations in Experiment 1 were intentionally created and selected to require visual information, and do not reflect the natural distribution of manga translation errors. Therefore, detection rates in Experiment 1 should not be interpreted as operational error detection rates or representative values for manga translation errors as a whole.
\item
Experiment 2 compares TOJ with IAJ-full; therefore, observed differences cannot be attributed solely to image presence because IAJ-full also modifies the evaluation instructions.
\item
The 5 runs in Experiment 1 repeat semantic-match classification of fixed MQM outputs, rather than MQM judge execution under identical conditions. MQM judge execution used \texttt{temperature=0.1}, but lowering the temperature does not guarantee identical outputs under repeated execution. Therefore, observed differences among TOJ, IAJ-minimal, IAJ-full, and IAJ-blank may partly reflect single-run MQM-output variability, in addition to differences in image input and evaluation instructions.
\item
Semantic match evaluation relies on LLM judge precision.
While authors performed spot-checks, alignment against independent multi-annotator gold labels and systematic verification of meta-evaluator precision were not conducted.
Consequently, reported semantic match rates may contain meta-evaluator judgment noise.
\item
We did not validate the judges' error localization, categories, severity labels, or explanations against independent MQM annotations by experienced translation experts. Human verification in Experiment 2 assessed judge outputs rather than independently annotating translation errors. Thus, our findings concern detection of injected degradations and differences in judge outputs, and do not establish overall MQM accuracy.
\item
Common flag matching between TOJ and IAJ-full employs a rule-based heuristic based on explanation similarity, category, and severity, without human alignment validation.
Therefore, unique flag counts and coverage rates may reflect heuristic matching errors.
\end{itemize}

\section*{Ethical Considerations}
\paragraph{Content Safety.}
The evaluated work is a general-audience medical drama containing medical depictions (surgeries, death), but review of the 180 evaluated pages revealed no explicit hate speech, discrimination, or explicit violence.

\section*{Acknowledgments}
We thank Shuho Sato and Sato Manga Works Ltd. for making
\emph{Give My Regards to Black Jack} available under a generous reuse
policy, which enabled this study and the release of the accompanying dataset.
This work was partly supported
by JST Moonshot R\&D Grant Number JPMJMS2011-35 (fundamental research),
by JSPS KAKENHI Grant Number JP26H02511,
and
by JST BOOST Japan Grant Number JPMJBS2421.

\bibliography{custom}

\appendix

\section{Prompts}
\label{sec:appendix-prompts}

Complete templates for primary prompts used in this study are shown below.
In actual execution, placeholders \texttt{\{...\}} were populated at runtime.

\paragraph{Degraded Translation Candidate Generation Prompt}
To generate degraded translation candidates, page images, source text, and official translations were input with the following prompt.
Degradation types were substituted at runtime as \texttt{\{degradation\_types\}}.
At runtime, the full-page source and official target text were substituted into \texttt{\{source\_full\}} and \texttt{\{translation\_full\}}, respectively. \texttt{\{pairs\}} contained bubble-level source-target pairs.

\begin{quote}\footnotesize\raggedright
You are generating natural but visually contradictory degradation candidates for manga translation evaluation.\\
The source language is Japanese and the target language is English.\\
For each of these degradation types, return up to 3 candidates in descending order of plausibility: \texttt{\{degradation\_types\}}.\\
Each candidate must change exactly one segment if possible and should only be clearly wrong when considering the image.\\
Do not create candidates that are clearly wrong from the source text alone. Avoid obvious text-only mismatches such as changing explicit names, countries, numbers, occupations, or key terms that are directly stated in the source text.\\
Also avoid candidates that become clearly wrong after reading neighboring text segments on the same page.\\
Prefer subtle changes where the source text and translation can still look plausible in isolation, but the page image makes the error clear.\\
Aim for candidates that remain plausible to a text-only judge, while still being strong enough to force a vision-aware judge to name a concrete translation error.\\
Accepted patterns: omitted Japanese subjects resolved differently by the image; gender-neutral references resolved by the depicted character; short replies whose tone depends on expression; object mismatch should target ambiguous references or generic words whose concrete referent is only visible in the image.\\
Rejected patterns: replacing an explicit source word with another term; adding a new action or proposition not licensed by the source or neighboring text; changing speaker/addressee when the page text already fixes it; making object/action changes that are already wrong without the image.\\
Type-specific constraints: avoid speaker/addressee changes when neighboring text already makes the addressee clear; avoid strong interjections whose tone is obvious from text alone; avoid count changes or lexical narrowing that already create a clear text-only meaning difference; avoid action changes when the source text already states the action or physical state.\\
The degraded\_text must be the rewritten translation segment only, in the same target language as the original translation. Do not output Japanese source text.\\
Return metadata:\\
- edit\_description\\
- visual\_evidence\_note\\
Only return JSON.

\textless{}input\textgreater{}\\
\textless{}page\_source\_full\textgreater{}\\
\{source\_full\}\\
\textless{}/page\_source\_full\textgreater{}\\
\textless{}page\_translation\_full\textgreater{}\\
\{translation\_full\}\\
\textless{}/page\_translation\_full\textgreater{}\\
\{pairs\}\\
\textless{}/input\textgreater{}\\
\end{quote}

\paragraph{Common Prompt for MQM Judges}
In TOJ, \texttt{\{vision\_instruction\}} and \texttt{\{evaluation\_procedure\}} are left empty.
In IAJ-minimal, only page images are added. In IAJ-full, additional instructions described below are embedded in \texttt{\{vision\_instruction\}} and \texttt{\{evaluation\_procedure\}}. IAJ-blank uses the IAJ-full prompt with page images replaced by a blank image.
MQM judge execution used \texttt{temperature=0.1}.

\begin{quote}\footnotesize\raggedright
\sloppy
You are an annotator for the quality of manga translation.\\
Your task is to identify translation errors on a single manga page and assess the overall quality.

\{vision\_instruction\}\\
\{evaluation\_procedure\}\\

\textless{}input\textgreater{}\\
\textless{}source\_language\textgreater{}\allowbreak\{src\_lang\}\allowbreak\textless{}/source\_language\textgreater{}\\
\textless{}target\_language\textgreater{}\{tgt\_lang\}\textless{}/target\_language\textgreater{}\\
\textless{}page\_source\_full\textgreater{}\\
\{source\_full\}\\
\textless{}/page\_source\_full\textgreater{}\\
\textless{}page\_translation\_full\textgreater{}\\
\{translation\_full\}\\
\textless{}/page\_translation\_full\textgreater{}\\
\{pairs\}\\
\textless{}/input\textgreater{}

First read the page as connected discourse, then evaluate individual segments.\\
Evaluate each source/translation segment pair carefully.\\
Also consider consistency, coherence, and fluency across neighboring segments on the same page.

Use the following error categories:\\
- meaning\_accuracy: Errors where the translation conveys a meaning that is substantially different from the source. Use this category only when no more specific category better explains the error.\\
- omission\_or\_addition: Errors where the translation omits source information that should be conveyed, or adds meaning that is not supported by the source.\\
- reference\_and\_disambiguation: Errors where the translation incorrectly resolves, over-specifies, or fails to appropriately preserve underspecified reference in the source.\\
- speaker\_and\_dialogue\_consistency: Errors where the translation is inconsistent with who is speaking, who is being addressed, or how the utterance fits into the local dialogue context on the page.\\
- tone\_and\_emotion: Errors where the translation fails to match the emotional tone, intensity, interpersonal stance, or expressive force conveyed by the source and page context.\\
- action\_and\_state: Errors where the translation misrepresents an action, posture, physical state, ongoing event, or target of action.\\
- terminology\_and\_named\_entities: Errors involving incorrect, inconsistent, or contextually inappropriate translation of specialized terms, role labels, names, titles, organizations, locations, or other named entities.\\
- naturalness\_and\_register: Errors where the translation is grammatically awkward, stylistically unnatural, or mismatched in register, formality, or social tone for the local context. Do not penalize a translation merely for being colloquial, slangy, or non-literal if it naturally conveys the source nuance in context.\\
- colloquial\_and\_cultural\_expression: Errors where the translation fails to appropriately render colloquial, stylized, era-specific, socially marked, or culturally grounded expression. Natural slang, casual phrasing, and expressive wording are acceptable when they help preserve the source nuance rather than flatten it into overly formal or literal English.\\
- other: Errors that materially affect translation quality but cannot be adequately explained by any category above. Use sparingly.

For each error: identify the primary segment\_index, optionally context\_segment\_indices, error\_span, explanation, exactly one error\_category, severity (major/minor/neutral), and improvement string.

Severity guideline: major = meaning substantially distorted; minor = meaning mostly preserved but suboptimal; neutral = limited impact.

Scoring guideline: Start from 100; subtract more for major than minor/neutral; keep scoring internally consistent across pages; do not double-count cross-segment issues (count a cross-segment issue as one error, not multiple duplicated errors).

Important constraints:\\
- Only annotate errors that are supported by the source text and translation.\\
- Do not invent missing context that is not justified by the input.\\
- Do not over-penalize stylistic variation unless it is clearly unnatural, misleading, or contextually inappropriate.\\
- Prefer translations that preserve source nuance, voice, and interpersonal texture, not just grammar or dictionary-level literalness.\\
- If the error is an omission, set ``error\_span'': ``'' and describe the missing content in ``explanation''. For omission errors, set improvement to the missing phrase or sentence.\\
- If an issue mainly appears when reading neighboring segments together, anchor it to the most affected segment\_index, include the relevant neighbors in context\_segment\_indices, and explain the cross-segment context in explanation.\\
- Do not report the same underlying issue multiple times. Merge it into one error anchored to the most affected segment.\\
- If there is no error, return an empty errors array.

Output requirements:\\
- Respond with valid JSON only.\\
- Do not include any text before or after the JSON.\\
- Use plain text strings without Markdown.
\end{quote}

\paragraph{MQM Judge JSON Schema}
The following JSON schema was inserted at runtime.

\begin{quote}\scriptsize
\begin{verbatim}
{
  "$defs": {
    "MQMError": {
      "properties": {
        "segment_index": {
          "minimum": 0,
          "title": "Segment Index",
          "type": "integer"
        },
        "context_segment_indices": {
          "items": {
            "minimum": 0,
            "type": "integer"
          },
          "title": "Context Segment Indices",
          "type": "array"
        },
        "error_span": {
          "title": "Error Span",
          "type": "string"
        },
        "explanation": {
          "title": "Explanation",
          "type": "string"
        },
        "error_category": {
          "enum": [
            "meaning_accuracy",
            "omission_or_addition",
            "reference_and_disambiguation",
            "speaker_and_dialogue_consistency",
            "tone_and_emotion",
            "action_and_state",
            "terminology_and_named_entities",
            "naturalness_and_register",
            "colloquial_and_cultural_expression",
            "other"
          ],
          "title": "Error Category",
          "type": "string"
        },
        "severity": {
          "enum": [
            "major",
            "minor",
            "neutral"
          ],
          "title": "Severity",
          "type": "string"
        },
        "improvement": {
          "title": "Improvement",
          "type": "string"
        }
      },
      "required": [
        "segment_index",
        "error_span",
        "explanation",
        "error_category",
        "severity",
        "improvement"
      ],
      "title": "MQMError",
      "type": "object"
    }
  },
  "properties": {
    "errors": {
      "items": {
        "$ref": "#/$defs/MQMError"
      },
      "title": "Errors",
      "type": "array"
    },
    "quality_score": {
      "maximum": 100,
      "minimum": 0,
      "title": "Quality Score",
      "type": "integer"
    },
    "visual_caption": {
      "default": "",
      "title": "Visual Caption",
      "type": "string"
    }
  },
  "required": [
    "errors",
    "quality_score"
  ],
  "title": "MQMStructuredResponse",
  "type": "object"
}
\end{verbatim}
\end{quote}

\paragraph{Additional Instructions for IAJ-full and IAJ-blank}
In IAJ-full and IAJ-blank, \texttt{\{vision\_instruction\}} and \texttt{\{evaluation\_procedure\}} in the common prompt are populated as follows:

\begin{quote}\footnotesize\raggedright
[vision\_instruction]\\
 Evaluate the translation with reference to the original manga image. Understand the visual content of the page carefully and use it to assess whether the translation is consistent with the depicted scene and context. Before judging translation quality, first form a concise internal caption of the page. Then identify the visible facts that matter for translation evaluation (who is present, who is likely speaking or being addressed, what object or person a referring expression points to, what action\_or\_state/emotional stance is depicted). Include the concise caption in the JSON field visual\_caption.

[evaluation\_procedure]\\
- Step 1 (visual grounding): Form a concise internal caption and extract only translation-relevant visual facts.\\
- Step 2 (MQM evaluation): Evaluate translation quality using source text, translation text, page-level discourse, and the grounded facts from Step 1.
\end{quote}

\paragraph{Semantic Match Classification Prompt}
Judge output error lists and degradation intents were input to an LLM (\texttt{gpt-5.6-luna}) to classify semantic match as \texttt{verdict} $\in \{\texttt{same}, \texttt{related}, \texttt{different}\}$.
An instance was counted as detected when at least one reported error received \texttt{verdict=same}; the proportion of detected instances was aggregated as the semantic match rate.
\texttt{gpt-5.6-luna} was executed with \texttt{reasoning\_effort=medium}.

\begin{quote}\footnotesize\raggedright
You are judging whether MQM errors semantically catch an intended degradation.\\
Decide if MQM has caught the intended degradation.

Degradation:\\
degradation\_type: \{degradation\_type\}\\
edit\_description: \{edit\_description\}\\
visual\_evidence\_note: \{visual\_evidence\_note\}

MQM Output:\\
mqm\_errors: \{mqm\_errors\}

Choose verdict and return concise reason:\\
- same: MQM catches essentially the same issue.\\
- related: MQM points to a related issue but not the intended issue strongly enough.\\
- different: MQM misses the intended issue.
\end{quote}

\clearpage
\section{Dataset Statistics}
\label{sec:appendix-dataset-stats}

Dataset details constructed and used in this study are shown in Table~\ref{tab:appendix-dataset-stats}.

\begin{table}[!htbp]
\centering
\footnotesize
\begin{tabular}{@{}p{0.57\columnwidth}p{0.35\columnwidth}@{}}
\toprule
\textbf{Item} & \textbf{Value} \\
\midrule
\multicolumn{2}{p{0.90\columnwidth}}{\emph{Base Dataset}} \\
Target Manga Work & \emph{Give My Regards to Black Jack} Vol. 1 \\
Page Images (JA) & 196 \\
Page Images (EN) & 196 \\
Pages with Established Segment Alignment & 180 \\
Total Segment Count (180 pages) & 974 \\
Mean Segments per Page & 5.4 \\
\midrule
\multicolumn{2}{p{0.90\columnwidth}}{\emph{Experiment 1 (Degraded Translation Detection Benchmark)}} \\
Degraded Translation Instances & 75 \\
Base Unique Pages & 54 \\
\multicolumn{2}{p{0.90\columnwidth}}{\emph{Instance Breakdown by Category}} \\
\texttt{speaker\_or\_addressee} & 16 \\
\texttt{referent\_or\_object} & 27 \\
\texttt{tone\_emotion\_or\_style} & 26 \\
\texttt{action\_or\_state} & 6 \\
\bottomrule
\end{tabular}
\caption{Dataset statistics.}
\label{tab:appendix-dataset-stats}
\end{table}

\section{Data Availability}
\label{sec:data-release}
The following materials are publicly available in the Hugging Face dataset repository at \url{https://huggingface.co/datasets/joddy/gmr2bj_manga_mt_ja_en}:
\begin{itemize}
\setlength{\itemsep}{0pt}
\item Page IDs, page numbers, speech bubble segments, and Japanese-English parallel segment alignment tables for 180 pages.
\item 75 degraded translation instances contradicting manga image content used in Experiment 1 (degraded target text, degradation type, intent, target segment, visual cue metadata).
\end{itemize}
\paragraph{Data Copyright and Licensing.}
All images in this study originate from data for \emph{Give My Regards to Black Jack} officially distributed for download by Sato Manga Works Ltd., rather than from scans of physical books.
Published terms of use allow copying, public transmission, adaptation, and secondary use under specified conditions. Figures and released datasets explicitly state ``Give My Regards to Black Jack / SHUHO SATO.''
Because this permission is specific to this work, it cannot be directly applied to other manga titles. Usage will be reported to Sato Manga Works Ltd. within one month of publication.
After verifying license compliance, the page images are released as part of the dataset. Official distribution page: \url{https://densho810.com/free/}.

\section{Experiment 1 Paired Cluster Test}
\label{sec:appendix-exp1-significance}

In Experiment 1, the same 75 degraded translation instances were evaluated by both TOJ (no image) and IAJ-minimal (with page image). Because these 75 instances originate from 54 base pages, we aggregated differences by source page to avoid counting instances from the same page as independent data points. We computed cluster sign-flip tests and cluster bootstrap 95\% confidence intervals across these 54 page-level differences.

Because the 5 runs represent re-executions over identical inputs, they are not pooled as 375 independent observations. Rather than repeating mean and run-to-run range values from main text Table~\ref{tab:exp1-main}, test results per run are reported in Table~\ref{tab:exp1-paired-significance-compare}.

For reference, exact McNemar test $p$-values based on instance-level paired disagreement are also listed. Because McNemar's test does not account for intra-page correlation, cluster sign-flip tests on base pages serve as our primary inference method.

\begin{table*}[htbp]
\centering
\small
\setlength{\tabcolsep}{4pt}
\input{tables/en/experiment1_paired_significance_compare}
\caption{Significance test results for TOJ vs. IAJ-minimal in Experiment 1. Differences per run are percentage-point differences (IAJ-minimal $-$ TOJ). ``TOJ-only detected'' indicates instances detected only by TOJ; ``IAJ-min.-only detected'' indicates instances detected only by IAJ-minimal, aggregated across 75 paired degraded translation instances. McNemar $p$ provides exact instance-level paired reference values. Cluster $p$ represents sign-flip test $p$-values over 54 page-level clusters, and Cluster 95\% confidence interval (CI) reflects bootstrap confidence intervals resampling source pages.}
\label{tab:exp1-paired-significance-compare}
\end{table*}

\section{Experiment 1: Additional Representative Examples}
\label{sec:appendix-exp1}

\begin{figure*}[htbp]
\centering

\begin{minipage}[t]{0.29\textwidth}
  \vspace{0pt}\centering
  \includegraphics[width=\linewidth,height=5.2cm,keepaspectratio]{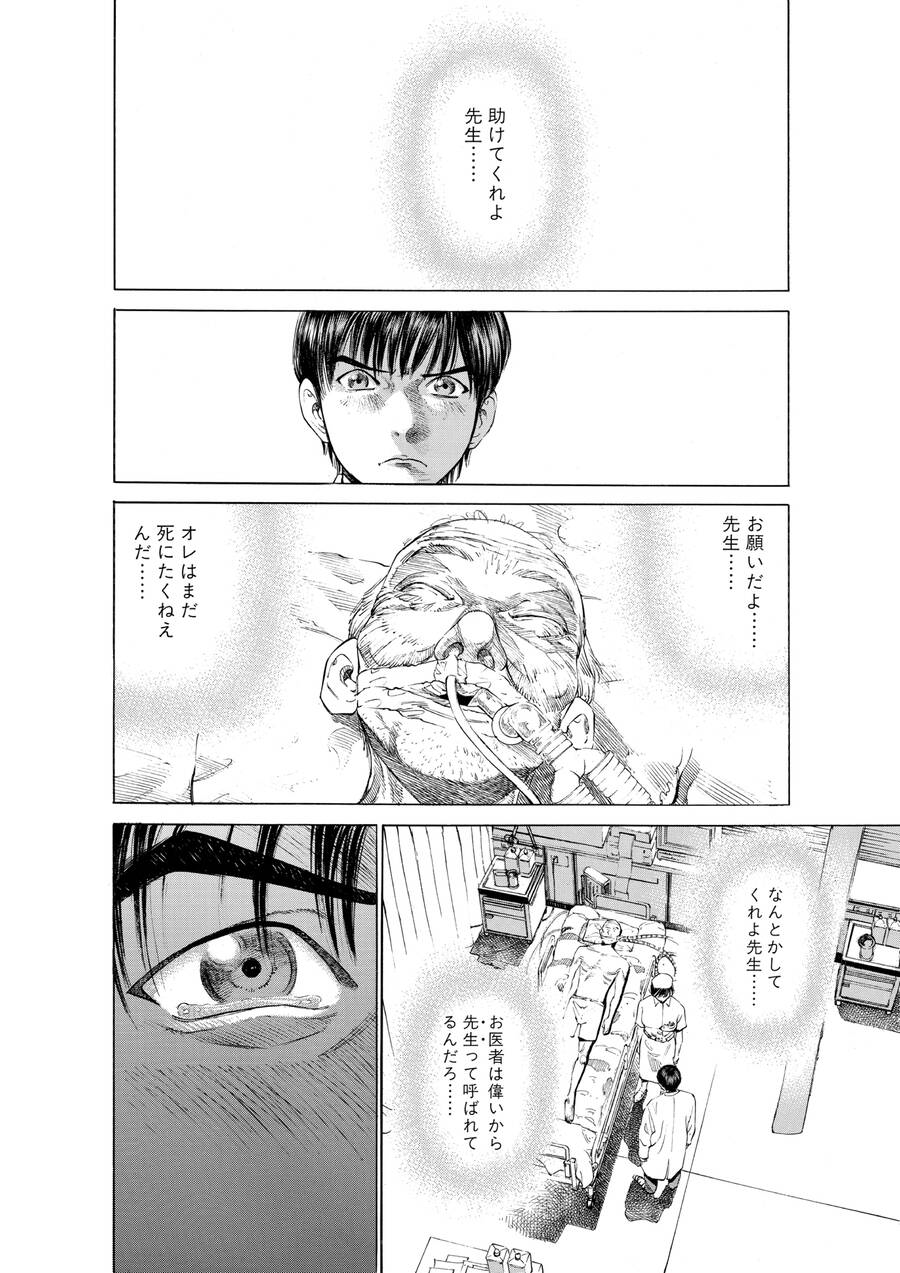}
  \caption*{\scriptsize bj\_vol1\_114 (JA)}
\end{minipage}\hfill
\begin{minipage}[t]{0.29\textwidth}
  \vspace{0pt}\centering
  \includegraphics[width=\linewidth,height=5.2cm,keepaspectratio]{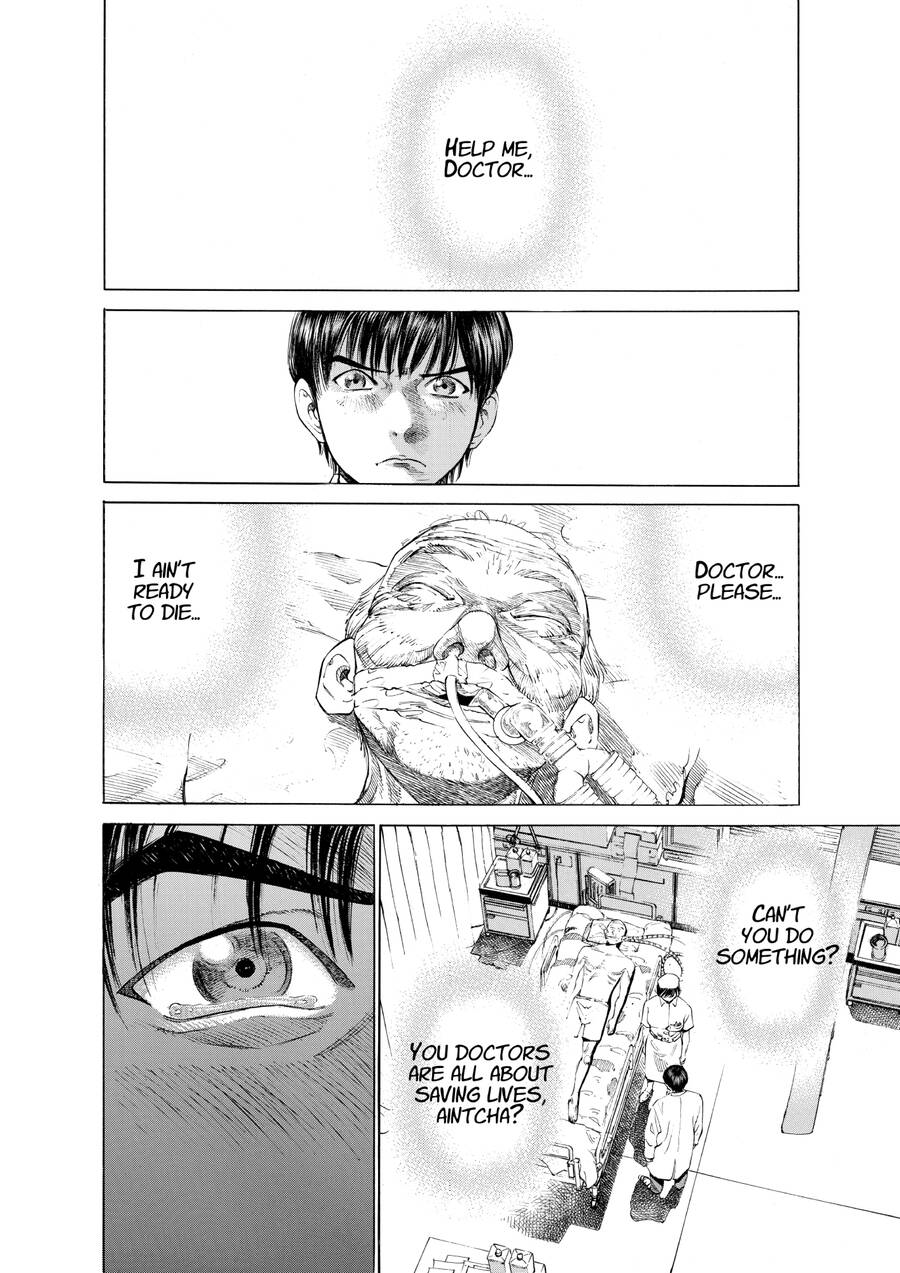}
  \caption*{\scriptsize bj\_vol1\_114 (EN)}
\end{minipage}\hfill
\begin{minipage}[t]{0.38\textwidth}
  \vspace{0pt}\scriptsize
  \textbf{Degraded segment}\\
  type: \texttt{tone\_emotion\_or\_style}\\
  JP ``助けてくれよ 先生…… 先生……''\\
  \texttt{\hspace*{1em}} EN (original) ``Help me, Doctor... Doctor...''\\
  \texttt{\hspace*{1em}} EN (degraded) ``Help me, Doctor! Doctor!''\\

  \textbf{Degradation intent}\\
  ``Converted the desperate, trailing-off internal thoughts (indicated by ellipses and fuzzy bubble borders) into assertive, forceful shouts by replacing ellipses with exclamation marks and using direct, clear phrasing.''\\

  \textbf{IAJ error explanation}\\
  ``The translation uses exclamation marks, conveying a forceful, shouting tone.\\
  However, the source uses ellipses (`……') and the visual context shows an intubated, dying patient, suggesting a weak, desperate plea or internal monologue.\\
  Additionally, `I AM NOT READY TO DIE' is a more composed phrasing than the raw `I don't want to die yet' (オレはまだ死にたくねえ), losing the directness and the nuance of `yet' (まだ).''
\end{minipage}

\vspace{0.8em}

\begin{minipage}[t]{0.29\textwidth}
  \vspace{0pt}\centering
  \includegraphics[width=\linewidth,height=5.2cm,keepaspectratio]{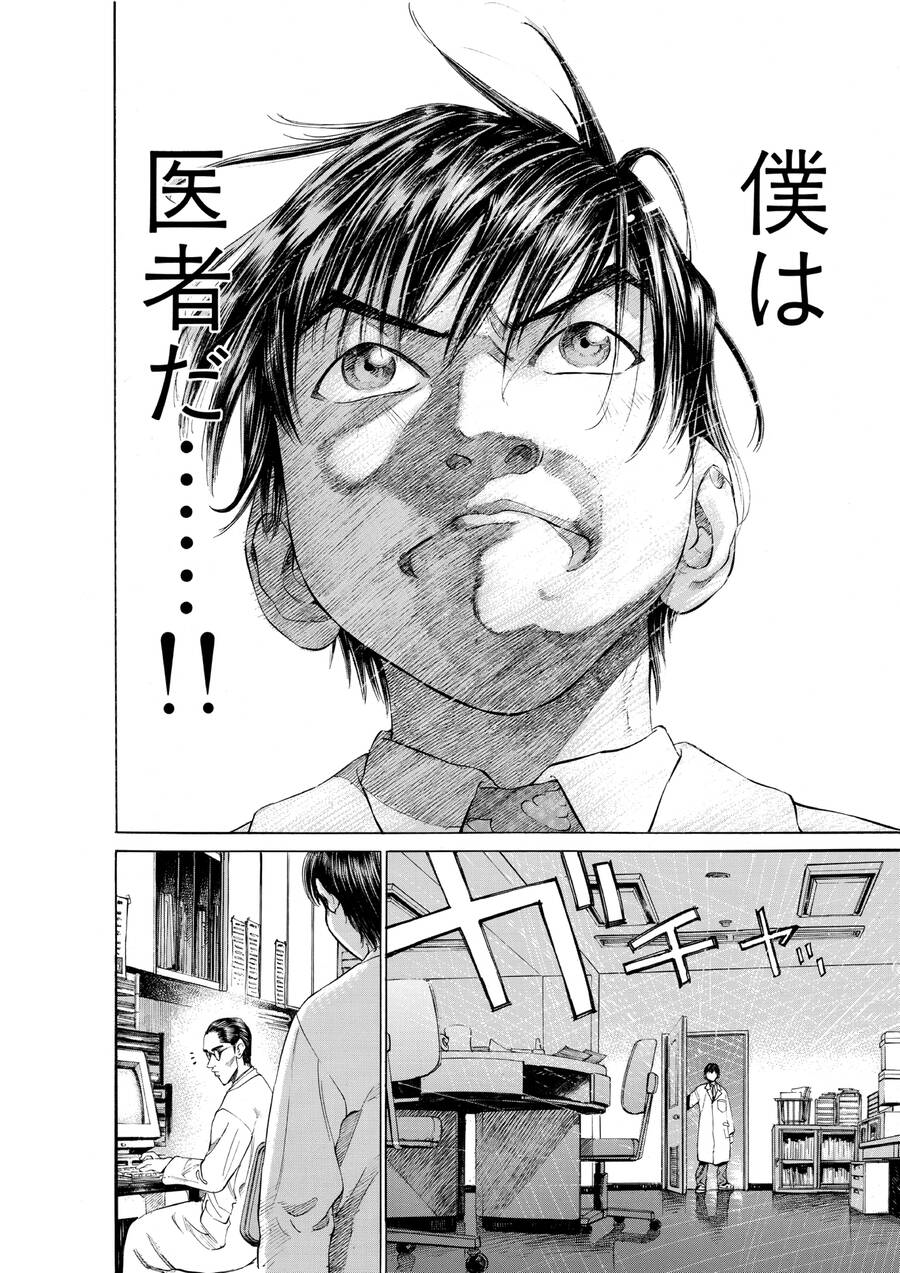}
  \caption*{\scriptsize bj\_vol1\_116 (JA)}
\end{minipage}\hfill
\begin{minipage}[t]{0.29\textwidth}
  \vspace{0pt}\centering
  \includegraphics[width=\linewidth,height=5.2cm,keepaspectratio]{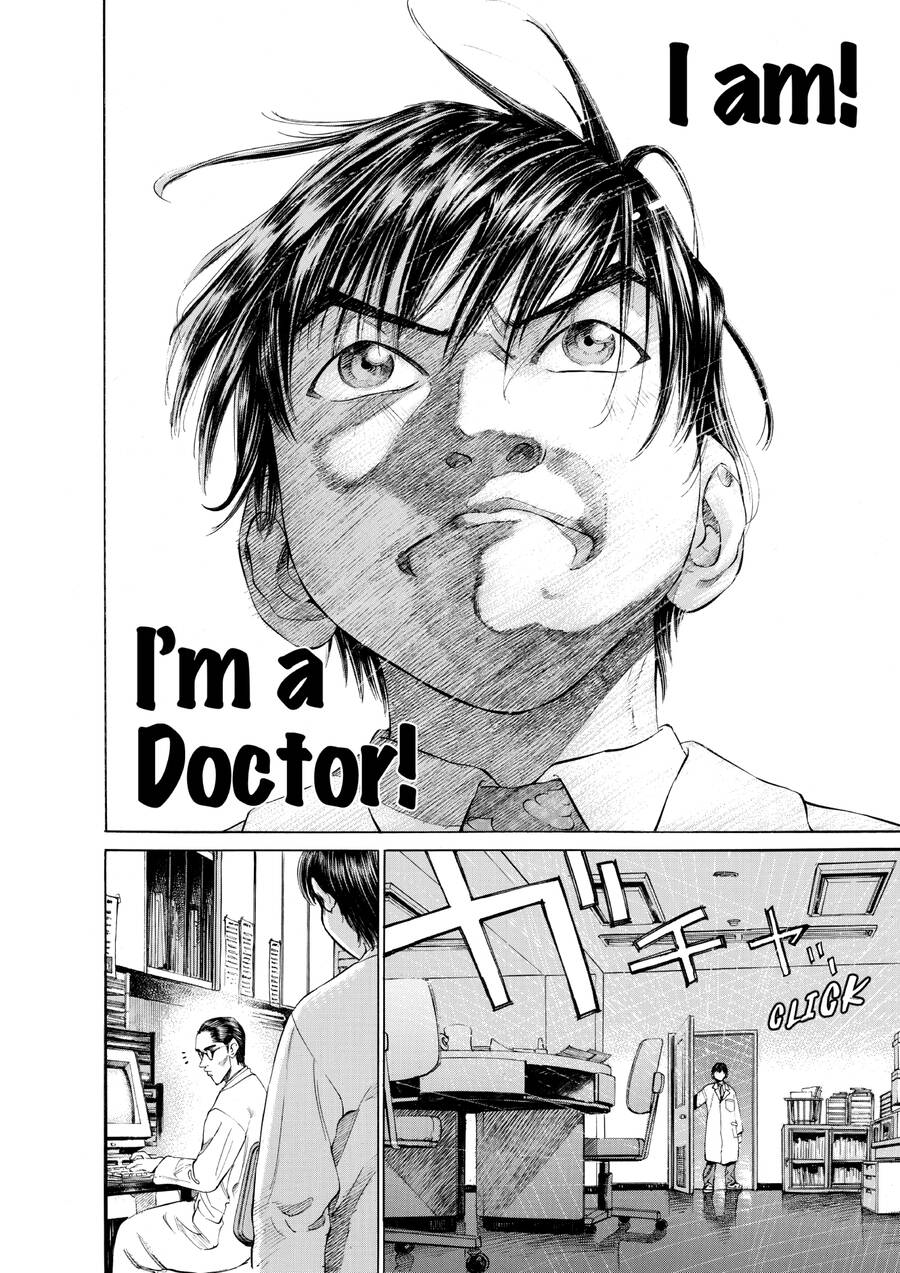}
  \caption*{\scriptsize bj\_vol1\_116 (EN)}
\end{minipage}\hfill
\begin{minipage}[t]{0.38\textwidth}
  \vspace{0pt}\scriptsize
  \textbf{Degraded segment}\\
  type: \texttt{tone\_emotion\_or\_style}\\
  JP ``僕は 医者だ・・・・・・！！'' \\
  \texttt{\hspace*{1em}} EN (original) ``I am! I'm a Doctor!''\\
  \texttt{\hspace*{1em}} EN (degraded) ``I... I'm a doctor...!!''\\

  \textbf{Degradation intent}\\
  ``Changed the tone from a fierce, defiant declaration of identity to a hesitant, fearful-sounding stutter.''\\

  \textbf{IAJ error explanation}\\
  ``The repetition of `I' followed by `I'm' creates a stuttering effect that suggests hesitation or uncertainty.\\
  This contradicts the character's extremely determined and resolute facial expression.\\
  The Japanese text uses a space for a dramatic pause between the subject and predicate, which is better rendered without repeating the pronoun to maintain the intensity of the declaration.''
\end{minipage}

\caption{Representative examples (\texttt{bj\_vol1\_114}, \texttt{bj\_vol1\_116}) of degraded translations contradicting manga image content in Experiment 1. Format matches Figure~\ref{fig:exp1-pages-main} (JA/EN page images + degraded segment + degradation intent + IAJ-full error explanation). The EN page images are shown for illustration only and were not provided to IAJ. \textit{Give My Regards to Black Jack / SHUHO SATO}}
\label{fig:exp1-pages}
\end{figure*}

Figure~\ref{fig:exp1-pages} presents 2 additional representative degraded translations contradicting manga image content: changing to forceful speech contradicting character state (\texttt{bj\_vol1\_114}) and changing to hesitant speech contradicting resolute determination (\texttt{bj\_vol1\_116}).

\section{Annotator Information}
\label{sec:appendix-annotators}
\paragraph{Execution Summary}
Six annotators evaluated the same set of 40 pages and completed post-task feedback.
The 6 annotators were researchers belonging to the same laboratory as the primary author (non-collaborators).
Compensation was paid equivalent to an hourly rate of 2,000 JPY.
Completing 40 pages required approximately 1.5 to 2 hours per annotator.

\begin{figure*}[htbp]
\centering
\includegraphics[width=\linewidth,keepaspectratio]{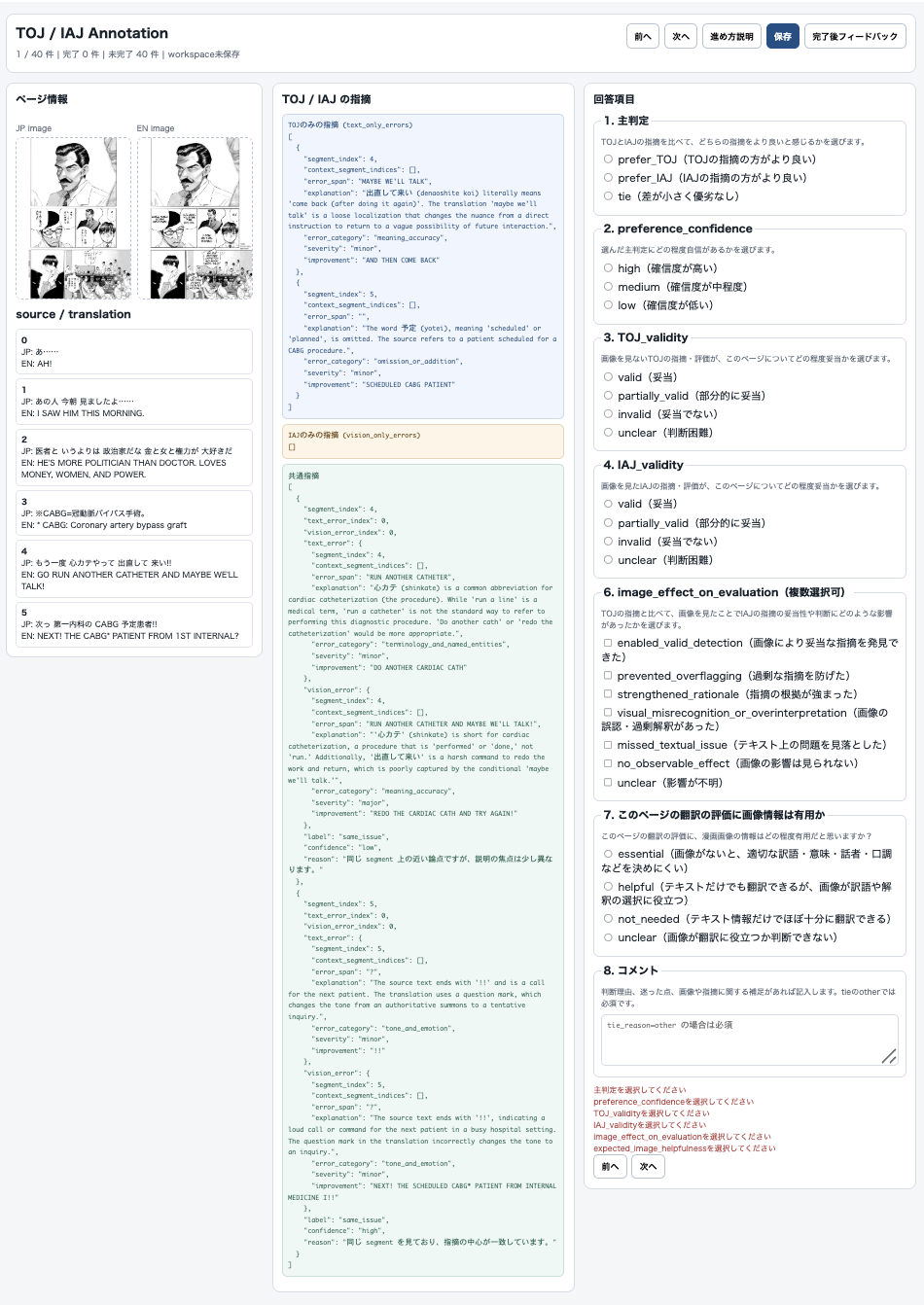}
\caption{Annotation UI screenshot. Left panel: JA/EN page images and comparison of source and translation segments. Middle panel: TOJ/IAJ flags. Right panel: annotation fields including main judgment (\texttt{prefer\_IAJ}/\texttt{prefer\_TOJ}/\texttt{tie}), confidence, page-level output validity, image effect, image usefulness, and comments. Images expand on click.}
\label{fig:annotation-ui}
\end{figure*}

\paragraph{Human Verification Task Items}
Rigid criteria were not enforced, allowing each annotator to judge independently.
A screenshot of the annotation UI is shown in Figure~\ref{fig:annotation-ui}.
Using identical UIs, annotators entered the following items per page:
\begin{itemize}
\item \textbf{Main Judgment}: Select which judge's output felt superior overall.
\begin{itemize}
\item \texttt{prefer\_TOJ}: TOJ flags were superior.
\item \texttt{prefer\_IAJ}: IAJ flags were superior.
\item \texttt{tie}: Difference was small or ranking was difficult. Must answer \texttt{tie\_reason}.
\end{itemize}
\item \textbf{Confidence}: Select confidence level in main judgment.
\begin{itemize}
\item \texttt{high}: High confidence in main judgment.
\item \texttt{medium}: Medium confidence in main judgment.
\item \texttt{low}: Low confidence; judgment felt unstable.
\end{itemize}
\item \textbf{Page-Level Output Validity}: Rate the overall validity of the complete TOJ and IAJ flag sets for the page independently on a 4-point scale.
\begin{itemize}
\item \texttt{valid}: The complete flag set is valid overall.
\item \texttt{partially\_valid}: The flag set is directionally valid but needs correction.
\item \texttt{invalid}: The complete flag set is invalid.
\item \texttt{unclear}: Cannot judge due to insufficient information.
\end{itemize}
\item \textbf{Image Effect}: \texttt{image\_effect\_on\_evaluation} (multiple-choice). Select how viewing images affected IAJ flags compared to TOJ.
\begin{itemize}
\item \texttt{enabled\_valid\_detection}: Discovered valid flag via image.
\item \texttt{prevented\_overflagging}: Prevented overflagging via image.
\item \texttt{strengthened\_rationale}: Strengthened flag rationale via image.
\item \texttt{visual\_misrecognition\_or\_}\allowbreak\texttt{overinterpretation}: Visual misrecognition or overinterpretation.
\item \texttt{missed\_textual\_issue}: Missed text issue due to viewing image.
\item \texttt{no\_observable\_effect}: No observable image effect (mutually exclusive).
\item \texttt{unclear}: Cannot judge image effect (mutually exclusive).
\end{itemize}
\item \textbf{Image Information Usefulness}: \texttt{expected\_image\_helpfulness} (single-choice). Rate how useful manga images were for evaluating translation quality on this page.
\begin{itemize}
\item \texttt{essential}: Difficult to judge proper wording/meaning/speaker/tone without image.
\item \texttt{helpful}: Text alone suffices, but image aids translation choices.
\item \texttt{not\_needed}: Text alone is sufficient.
\item \texttt{unclear}: Cannot judge image usefulness.
\end{itemize}
This item represents potential image usefulness as perceived by evaluators.
\item \textbf{Tie Reason}: Single-choice reason when selecting \texttt{tie}.
\begin{itemize}
\item \texttt{both\_valid}: Both TOJ and IAJ flags are valid.
\item \texttt{both\_partially\_valid}: Both flags are partially valid.
\item \texttt{both\_invalid}: Both flags are invalid.
\item \texttt{difference\_too\_small}: Output difference is too small to rank.
\item \texttt{rubric\_dependent}: Judgment depends on which evaluation criterion is prioritized.
\item \texttt{other}: Other reason (requires text comment).
\end{itemize}
\end{itemize}

\paragraph{Post-Task Feedback Items}
After completing 40 pages, annotators answered post-task feedback items on 5-point Likert scales (1 = strongly disagree, 5 = strongly agree):
\begin{itemize}
\item \textbf{5-Point Likert Items}:
\begin{itemize}
\item \texttt{overall\_difficulty}: Comparing TOJ and IAJ flags to judge which was better was difficult.
\item \texttt{rubric\_sufficiency}: Procedure explanations and item descriptions were sufficient to judge consistently.
\item \texttt{priority\_clarity}: Priority rules were clear when source fidelity, naturalness, and visual fit conflicted.
\item \texttt{preference\_label\_clarity}: Criteria for selecting \texttt{prefer\_TOJ}, \texttt{prefer\_IAJ}, and \texttt{tie} were clear.
\item \texttt{example\_helpfulness}: Concrete label examples or boundary cases would have aided decision-making.
\end{itemize}
\item \textbf{Hardest Annotation Field} (\texttt{hardest\_annotation\_field}, single-choice):
\begin{itemize}
\item \texttt{overall\_preference}: Main judgment.
\item \texttt{preference\_confidence}: Judgment confidence.
\item \texttt{TOJ\_validity}: TOJ page-level output validity rating.
\item \texttt{IAJ\_validity}: IAJ page-level output validity rating.
\item \texttt{tie\_reason}: Reason for selecting tie.
\item \texttt{image\_effect\_on\_evaluation}: Image effect on IAJ flags.
\item \texttt{expected\_image\_helpfulness}: Image usefulness for translation evaluation.
\item \texttt{none}: No item was particularly difficult.
\end{itemize}
\item \textbf{Reasons for Difficulty} (\texttt{difficulty\_reasons}, multiple-choice):
\begin{itemize}
\item \texttt{small\_difference}: Output difference between TOJ and IAJ was small.
\item \texttt{both\_valid}: Both flags were valid.
\item \texttt{both\_problematic}: Both flags had flaws.
\item \texttt{fidelity\_vs\_naturalness}: Conflict between source fidelity and target naturalness.
\item \texttt{text\_vs\_visual\_fit}: Difficulty evaluating text vs. visual fit.
\item \texttt{severity\_threshold}: Severity thresholds were difficult to calibrate.
\item \texttt{insufficient\_context}: Context was insufficient for judgment.
\item \texttt{complex\_judge\_outputs}: Judge output content was complex.
\item \texttt{unclear\_label\_boundaries}: Label boundaries were unclear.
\item \texttt{image\_quality}: Image quality hindered judgment.
\item \texttt{ui\_issue}: UI display or operation issues.
\item \texttt{other}: Other reasons.
\end{itemize}
\item \textbf{Most Prioritized Criterion} (\texttt{prioritized\_criterion}, single-choice):
\begin{itemize}
\item \texttt{source\_fidelity}: Source fidelity.
\item \texttt{translation\_naturalness}: English translation naturalness.
\item \texttt{visual\_consistency}: Consistency with manga image.
\item \texttt{error\_severity}: Flagged error severity.
\item \texttt{overall\_balance}: Overall balance across criteria.
\item \texttt{case\_by\_case}: Evaluated on a case-by-case basis.
\item \texttt{unclear}: Could not set consistent criteria.
\end{itemize}
\item \textbf{Desired Guidance Types} (\texttt{requested\_guidance}, multiple-choice):
\begin{itemize}
\item \texttt{label\_examples}: Concrete examples for each label.
\item \texttt{boundary\_examples}: Examples of boundary cases.
\item \texttt{tie\_examples}: Concrete examples of tie cases.
\item \texttt{priority\_rules}: Priority rules when criteria conflict.
\item \texttt{severity\_rules}: Guidelines for severity assessment.
\item \texttt{image\_effect\_examples}: Concrete examples of image effects.
\item \texttt{full\_walkthrough}: End-to-end task walkthrough.
\item \texttt{more\_context}: Additional context.
\item \texttt{none}: No additional guidance needed.
\end{itemize}
\item \textbf{Free Comments} (\texttt{free\_comment}, optional): Free-text input for hesitation details, UI feedback, or requested explanations.
\end{itemize}

\section{Automatic Matching and Score Aggregation}
\label{sec:appendix-base-rate}

To detail the automatic output differences reported in \S\ref{sec:exp2}, we aggregated TOJ / IAJ-full flag counts on official translations across 180 pages (main text Table~\ref{tab:exp2-base-rate-main}).

\paragraph{Common Flag Matching Procedure}
Common flags in this paper are established by \textbf{rule-based automatic matching rather than human annotation}.
Specifically, candidate TOJ/IAJ error pairs within the same \texttt{segment\_index} are created, taking the maximum of explanation token Jaccard similarity and \texttt{SequenceMatcher} string similarity as the base score.
We add $+0.15$ for matching \texttt{error\_category} and $+0.05$ for matching \texttt{severity}, capping the score at 1.0. For the \texttt{SequenceMatcher} component, explanations are normalized with NFKC normalization, lowercasing, and removal of non-word characters; therefore, exact matches after this normalization already have a base similarity of 1.0.
Candidates are sorted by score descending, executing 1-to-1 greedy matching while excluding previously matched errors.
Pairs with \texttt{score < 0.45} are rejected without an exact-match exception. Exact normalized explanations still require 1-to-1 pairing within the same segment and do not guarantee human semantic identity.
Accepted pairs are stored as \texttt{same\_issue} in \texttt{common\_errors}.
Unmatched flags are output as \texttt{text\_only\_errors} or \texttt{vision\_only\_errors}.
Restricting matching candidates to identical segments and combining string similarity with \texttt{error\_category} and \texttt{severity} allows this matching to serve as a practical classification separating common and unique flags.
However, common classification does not guarantee strict semantic identity or human ground truth, and inter-annotator agreement (IAA) for common matching was not separately calculated.
Accordingly, TOJ-only / IAJ-only / common aggregations in \S\ref{sec:exp2} should be interpreted as diagnostic statistics comparing judge output trends rather than human gold labels.

\subsection{Model-Reported Scores and Auxiliary Score Deficits}
The \texttt{quality\_score} in this study is an auxiliary model-reported score directly output by judges alongside error lists, rather than a standard MQM quality score calculated from fixed severity weights.
Prompts instructed judges to start from 100, penalize \texttt{major} more than \texttt{minor} or \texttt{neutral}, and avoid double-counting, but did not fix numerical penalty amounts per severity or aggregate formulas.
Therefore, this score should not be interpreted as a raw MQM penalty sum or a metric mechanically recomputed from error counts.

For comparison, we define the auxiliary score deficit from model-reported scores as \(D^{\mathrm{judge}}_p = 100 - S_p\) per page.
Mean score deficits reported in Table~\ref{tab:exp2-base-rate-main} and derived from the mean scores in Table~\ref{tab:exp2-robustness} represent \(D^{\mathrm{judge}}_p\) averaged across 180 pages.

\subsection{Fixed-Weight Scoring Based on MQM Severity and Normalized Penalties}
In addition to \texttt{quality\_score}, we conducted an auxiliary analysis aggregating reported severity labels with fixed weights.
Letting $n_{p,\sigma}$ be the count of flags with severity $\sigma$ on page $p$, we assign weights of 0, 1, and 5 to neutral, minor, and major severities, respectively, defining raw MQM penalty as:
\begin{equation}
D^{\mathrm{MQM}}_p = 0n_{p,\mathrm{neutral}} + 1n_{p,\mathrm{minor}} + 5n_{p,\mathrm{major}}
\end{equation}
To account for text length variation across pages, we normalize penalty by official target word count $W_p$:
\begin{equation}
E_p^{\mathrm{word}} = 100\frac{D^{\mathrm{MQM}}_p}{W_p}
\end{equation}
representing penalty points per 100 target words. The primary aggregation is computed as a dataset micro-average across all 180 pages:
\begin{equation}
E_{\mathrm{micro}}^{\mathrm{word}} = 100\frac{\sum_p D^{\mathrm{MQM}}_p}{\sum_p W_p}
\end{equation}
where lower values indicate smaller penalties.

Across all 3 models (Gemini, GPT-5.4-mini, Qwen3-VL), IAJ-full yielded lower normalized penalty-point values than TOJ. The difference from TOJ to IAJ-full (IAJ-full $-$ TOJ) was $-1.23$, $-1.05$, and $-3.93$ penalty points per 100 target words, respectively. Page counts where IAJ-full penalty was lower, TOJ penalty was lower, or tied were 62/21/97 for Gemini, 70/50/60 for GPT-5.4-mini, and 92/42/46 for Qwen3-VL. Sensitivity checks across 4 different denominator choices (per 100 target words, per 10 segments, per 100 source characters, and per 100 target words excluding short target pages) consistently preserved directional gains across all 3 models. However, this does not directly validate individual flag accuracy or human correlation superiority.

\subsection{Correlation Between Severity-Weighted Auxiliary Penalties and \texttt{quality\_score}}
To examine the alignment between auxiliary penalties and \texttt{quality\_score}, we calculated page-level correlations between \texttt{quality\_score} and sign-flipped normalized MQM penalties ($-E_p^{\mathrm{word}}$). Results are shown in Table~\ref{tab:exp2-mqm-quality-correlation}.

\begin{table*}[t]
\centering
\footnotesize
\setlength{\tabcolsep}{5pt}
\begin{tabular}{@{}llrr@{}}
\toprule
\textbf{Model} & \textbf{Condition} & \textbf{Pearson $r$} & \textbf{Spearman $\rho$} \\
\midrule
\texttt{gemini-3-flash-preview} (Main) & TOJ & 0.402 & 0.762 \\
\texttt{gemini-3-flash-preview} (Main) & IAJ-full & 0.348 & 0.717 \\
\texttt{gpt-5.4-mini} (Auxiliary) & TOJ & 0.557 & 0.873 \\
\texttt{gpt-5.4-mini} (Auxiliary) & IAJ-full & 0.752 & 0.881 \\
\texttt{qwen3-vl-32b-instruct} (Auxiliary) & TOJ & 0.467 & 0.803 \\
\texttt{qwen3-vl-32b-instruct} (Auxiliary) & IAJ-full & 0.391 & 0.693 \\
\midrule
3 Models, 2 Conditions (Pooled) & All Conditions & 0.492 & 0.817 \\
\bottomrule
\end{tabular}
\caption{Page-level correlation between \texttt{quality\_score} and sign-flipped normalized MQM penalty ($-E_p^{\mathrm{word}}$). Each model and condition has $n=180$; pooled values have $n=1080$.}
\label{tab:exp2-mqm-quality-correlation}
\end{table*}

Pooled across all conditions, Pearson $r=0.492$ and Spearman $\rho=0.817$, showing general ranking alignment between model-reported \texttt{quality\_score} and severity-weighted auxiliary penalties.
However, Pearson correlations varied by model and condition ($0.348$--$0.752$), indicating they cannot be treated as identical scale metrics. Removing text volume effects via normalization also weakened Pearson correlation relative to raw penalty sums.

Comparing improvement magnitudes from TOJ to IAJ-full between $\Delta S_p = S_{p,\mathrm{IAJ}} - S_{p,\mathrm{TOJ}}$ and $G_p = E_{p,\mathrm{TOJ}}^{\mathrm{word}} - E_{p,\mathrm{IAJ}}^{\mathrm{word}}$, Pearson/Spearman correlations were $0.779/0.783$ for \texttt{gemini-3-flash-preview}, $0.586/0.731$ for \texttt{gpt-5.4-mini}, and $0.726/0.820$ for \texttt{qwen3-vl-32b-instruct} (pooled 540 pairs: $0.664/0.773$).
Thus, both metrics demonstrate moderate-to-strong alignment in improvement direction, though this correlation reflects convergent metric behavior rather than human ground-truth alignment.

\section{Experiment 2 Model-Reported Score Results}
\label{sec:appendix-exp2-robustness}

In Experiment 2, we conducted automatic MQM evaluation with the primary model, \texttt{gpt-5.4-mini}, and \texttt{qwen3-vl-32b-instruct} on 180 official translation pages. Table~\ref{tab:exp2-robustness} compares reported page scores (\texttt{quality\_score}). Across all three models, IAJ-full yielded higher mean model-reported scores and lower auxiliary score deficits than TOJ. The number of pages where IAJ-full scored higher than TOJ also exceeded the reverse. However, this is an auxiliary diagnostic showing judge output distribution shifts on official translations, not proof of page-level output validity or universal superiority.

\begin{table*}[t]
\centering
\footnotesize
\begin{tabular}{lrrrrrr}
\toprule
\textbf{Model} & \textbf{IAJ Higher} & \textbf{TOJ Higher} & \textbf{Tied} &
\textbf{Mean IAJ} & \textbf{Mean TOJ} & \textbf{Mean Diff} \\
\midrule
\texttt{gemini-3-flash-preview} & 81 (45.0\%) & 32 (17.8\%) & 67 (37.2\%) & 91.9 & 89.8 & +2.2 \\
\texttt{gpt-5.4-mini} & 90 (50.0\%) & 43 (23.9\%) & 47 (26.1\%) & 87.9 & 86.4 & +1.5 \\
\texttt{qwen3-vl-32b-instruct} & 102 (56.7\%) & 24 (13.3\%) & 54 (30.0\%) & 85.5 & 80.1 & +5.4 \\
\bottomrule
\end{tabular}
\caption{Experiment 2 score summary across the primary and auxiliary models. For 180 official translation pages, the table shows page counts where the IAJ-full model-reported score exceeded TOJ, vice versa, ties, and mean model-reported scores (\texttt{quality\_score}). Mean difference reflects average unrounded score difference (IAJ $-$ TOJ).}
\label{tab:exp2-robustness}
\end{table*}

Error volume aggregations for auxiliary models in the format of main text Table~\ref{tab:exp2-base-rate-main} are shown in Table~\ref{tab:exp2-robustness-error-volume}.
In \texttt{gpt-5.4-mini}, IAJ-full produced fewer total error flags than TOJ, and \texttt{qwen3-vl-32b-instruct} similarly yielded fewer total and major flags.
However, in Qwen, minor flags for IAJ-full exceeded TOJ, showing severity-level shifts vary by architecture.

\begin{table*}[t]
\centering
\footnotesize
\begin{tabular}{@{}lrrrr@{}}
\toprule
\textbf{Metric} & \textbf{gpt TOJ} & \textbf{gpt IAJ-full} &
\textbf{qwen TOJ} & \textbf{qwen IAJ-full} \\
\midrule
\multicolumn{5}{l}{\emph{All 180 Pages}} \\
Total error flags & 463 & 444 & 568 & 547 \\
Mean / page & 2.57 & 2.47 & 3.16 & 3.04 \\
Major error flags & 136 & 123 & 212 & 146 \\
Mean major / page & 0.76 & 0.68 & 1.18 & 0.81 \\
Minor error flags & 301 & 285 & 348 & 375 \\
Mean minor / page & 1.67 & 1.58 & 1.93 & 2.08 \\
Neutral error flags & 26 & 36 & 8 & 26 \\
Mean neutral / page & 0.14 & 0.20 & 0.04 & 0.14 \\
\midrule
\multicolumn{5}{l}{\emph{Page-Level Error Distribution}} \\
0-error pages & 24 (13.3\%) & 36 (20.0\%) & 11 (6.1\%) & 4 (2.2\%) \\
1+ error pages & 156 (86.7\%) & 144 (80.0\%) & 169 (93.9\%) & 176 (97.8\%) \\
0-major pages & 84 (46.7\%) & 92 (51.1\%) & 52 (28.9\%) & 85 (47.2\%) \\
0-error in both judges & \multicolumn{2}{c}{24 (13.3\%)} & \multicolumn{2}{c}{3 (1.7\%)} \\
\bottomrule
\end{tabular}
\caption{Error volume reported by auxiliary models on 180 official translation pages. Total error flags include \texttt{major}, \texttt{minor}, and \texttt{neutral}. gpt represents \texttt{gpt-5.4-mini}; qwen represents \texttt{qwen3-vl-32b-instruct}.}
\label{tab:exp2-robustness-error-volume}
\end{table*}

\section{Detailed TOJ Coverage in IAJ-full}
\label{sec:appendix-coverage}

\paragraph{TOJ Flag Coverage by IAJ-full}
Regarding coverage mentioned in \S\ref{sec:exp2}, mathematical definition, target set, micro-average, and page-level distributions are detailed in Table~\ref{tab:coverage-distribution}.
For each page $p$, let $\mathrm{common}_p$ and $\mathrm{TOJ\text{-}only}_p$ denote the numbers of retained and non-retained TOJ flags, respectively. Page-level coverage is defined as:
\begin{equation}
c_p = \frac{\mathrm{common}_p}{\mathrm{TOJ\text{-}only}_p + \mathrm{common}_p}.
\end{equation}
The dataset-level micro-average is defined as:
\begin{equation}
c_{\mathrm{micro}} = \frac{\sum_p \mathrm{common}_p}{\sum_p(\mathrm{TOJ\text{-}only}_p + \mathrm{common}_p)}.
\end{equation}
We additionally report the page-level macro-average and median of $c_p$.
Aggregated over 160 pages with at least 1 TOJ error flag (excluding 20 pages with 0 TOJ flags), nearly 30\% (45 pages; 28.1\%) exhibited 0\% coverage (IAJ retained zero TOJ flags). This indicates that evaluation perspectives between TOJ and IAJ diverge significantly on select pages.

\begin{table*}[htbp]
\centering
\small
\begin{tabular}{p{0.60\textwidth}r}
\toprule
\textbf{Metric / Page Group} & \textbf{Value / Pages} \\
\midrule
Micro-Average Coverage (166 / 318 flags) & 52.2\% \\
Page-Level Macro-Average Coverage (n=160) & 53.1\% \\
Page-Level Median Coverage & 50.0\% \\
\midrule
100\% Coverage (Fully Retained) Pages & 54 pages (33.8\%) \\
Partial Coverage ($0 < c_p < 1$) Pages & 61 pages (38.1\%) \\
0\% Coverage (Fully Suppressed) Pages & 45 pages (28.1\%) \\
\bottomrule
\end{tabular}
\caption{Coverage of TOJ error flags by IAJ-full and page-level distributions.}
\label{tab:coverage-distribution}
\end{table*}

\end{document}

%% file: tables/en/experiment1_factorial_ablation.tex
\begin{tabular}{lccrrr}
\toprule
\textbf{Condition} & \textbf{Page Image} & \textbf{Grounding} & \textbf{Mean (\%)} & \textbf{5-run Range (\%)} & \textbf{Mean flags} \\
\midrule
TOJ    & None        & None & 57.9\% & 56.0--58.7\% & 1.11 \\
IAJ-blank   & Blank image & Yes & 52.0\% & 50.7--53.3\% & 1.03 \\
IAJ-minimal & Actual page & None & 81.6\% & 80.0--82.7\% & 1.40 \\
IAJ-full    & Actual page & Yes & 80.5\% & 78.7--82.7\% & 1.20 \\
\bottomrule
\end{tabular}

%% file: tables/en/experiment1_paired_significance_compare.tex
\begin{tabular}{lrrrrrrrr}
\toprule
\textbf{Run} & \textbf{TOJ (\%)} & \textbf{IAJ-min. (\%)} & \textbf{Diff (pp)} &
\textbf{TOJ-only} & \textbf{IAJ-min.-only} & \textbf{McNemar $p$} &
\textbf{Cluster $p$} & \textbf{Cluster 95\% CI} \\
\midrule
1 & 58.7\% & 81.3\% & +22.7 & 2 & 19 & 0.000221 & 0.000740 & [+11.0, +36.1] \\
2 & 58.7\% & 80.0\% & +21.3 & 2 & 18 & 0.000403 & 0.001700 & [+8.3, +35.1] \\
3 & 56.0\% & 81.3\% & +25.3 & 1 & 20 & $<0.0001$ & $<0.0001$ & [+13.3, +38.6] \\
4 & 57.3\% & 82.7\% & +25.3 & 1 & 20 & $<0.0001$ & $<0.0001$ & [+13.9, +37.8] \\
5 & 58.7\% & 82.7\% & +24.0 & 1 & 19 & $<0.0001$ & 0.000390 & [+11.5, +37.7] \\
\bottomrule
\end{tabular}